\documentclass[10pt,twocolumn,letterpaper]{article}

\usepackage[pagenumbers]{cvpr} % To force page numbers, e.g. for an arXiv version

\usepackage[dvipsnames]{xcolor}

\definecolor{cvprblue}{rgb}{0.21,0.49,0.74}
\usepackage[pagebackref,breaklinks,colorlinks,citecolor=cvprblue]{hyperref}
\usepackage{lineno,hyperref,lettrine}
\usepackage{graphicx,comment}
\usepackage{amsmath,amssymb,booktabs,multirow,bm} % define this before the line numbering.
\usepackage{xcolor,float,tikz,pgfplots,subcaption}
\usepackage[normalem]{ulem}
\def\paperID{7960} % *** Enter the Paper ID here
\def\confName{CVPR}
\def\confYear{2024}

\title{CSA-Net: Channel-wise Spatially Autocorrelated\\ Attention Networks}

\author{Nick Nikzad\\
Institute for Integrated and Intelligent Systems\\
Griffith University, Australia\\
{\tt\small n.nikzaddehaji@griffith.edu.au}
\and
Yongsheng Gao\\
Institute for Integrated and Intelligent Systems\\
Griffith University, Australia\\
{\tt\small yongsheng.gao@griffith.edu.au}
\and
Jun Zhou\\
Institute for Integrated and Intelligent Systems\\
Griffith University, Australia\\
{\tt\small jun.zhou@griffith.edu.au}
}

\begin{document}
\maketitle
\begin{abstract}
In recent years, convolutional neural networks (CNNs) with channel-wise feature refining mechanisms have brought noticeable benefits to modelling channel dependencies. However, current attention paradigms fail to infer an optimal channel descriptor capable of simultaneously exploiting statistical and spatial relationships among feature maps. In this paper, to overcome this shortcoming, we present a novel channel-wise spatially autocorrelated (CSA) attention mechanism. Inspired by geographical analysis, the proposed CSA exploits the spatial relationships between channels of feature maps to produce an effective channel descriptor. To the best of our knowledge, this is the first time that the concept of geographical spatial analysis is utilized in deep CNNs. The proposed CSA imposes negligible learning parameters and light computational overhead to the deep model, making it a powerful yet efficient attention module of choice. We validate the effectiveness of the proposed CSA networks (CSA-Nets) through extensive experiments and analysis on ImageNet, and MS COCO benchmark datasets for image classification, object detection, and instance segmentation. The experimental results demonstrate that CSA-Nets are able to consistently achieve competitive performance and superior generalization than several state-of-the-art attention-based CNNs over different benchmark tasks and datasets.
\end{abstract}    
\section{Introduction}
\label{sec:intro}

Over the last decade, convolutional neural networks (CNNs) have shown outstanding performance on a variety of computer vision applications~\cite{ImageNet}. Inspired by the biological process of the human visual system, they produce discriminative and representative features from raw input signals without any prior manipulation. In general, convolutional neural networks (CNNs) are constructed by stacking a series of convolution layers, non-linear activation and down-sampling functions. All these architectures are capable of effectively capturing hierarchical patterns along all the input channels. While most early researches~\cite{VGG,he2016identity,ResNetXt,GoogleNet} mainly focused on improving the joint encoding of spatial and channel information, some works attempted to exploit and model information carried by inter-dependencies among channels. Methods like squeeze and excitation networks (SENets)~\cite{senet}, bottleneck attention module (BAM)~\cite{BAM}, convolutional block attention module (CBAM)~\cite{CBAM}, global-and-local (GALA)~\cite{GALA}, and dense-and-implicit attention network (DIANet)~\cite{dianet} adopt channel-wise gating mechanisms to learn nonlinear synergies between channels in order to re-calibrate the feature map.

\begin{figure}[t]
\centering
	\centering
	\includegraphics[trim=0.5cm .5cm 0.2cm 0.0cm, clip, width=0.5\textwidth,scale=0.60]{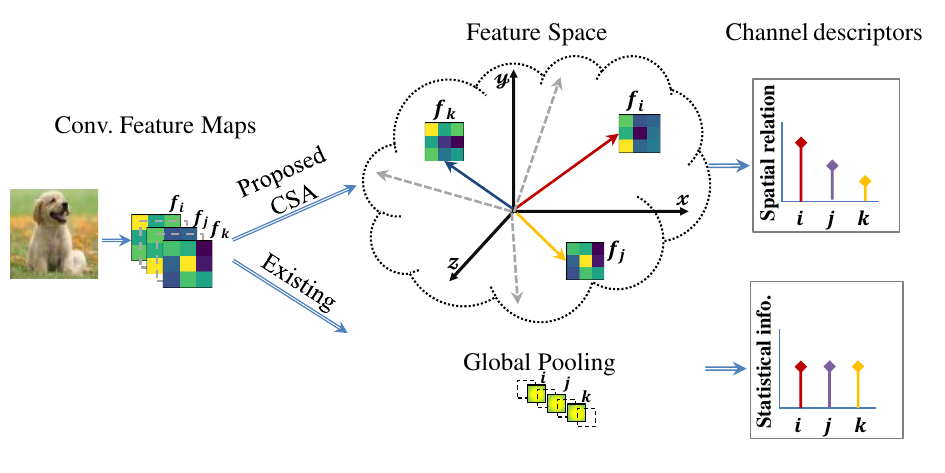}%
	\caption{Conceptual comparison of the generic channel descriptors used by most of existing attention mechanisms~\cite{senet,GENet,CBAM,GALA,dianet} and the proposed one. Any image passing a convolutional layer is represented by a set of feature maps (Assume $f_{i}$,$f_{j}$, and $f_{k} \in \mathbb{R}^3$). Due to the data dependency of statistical pooling operators, feature maps show similar statistical properties while their spatial relationships are distinctive.}
	\label{fig:SE_issue}
\end{figure}
Despite the advantage of existing attention mechanisms~\cite{senet,CBAM,GALA,dianet,ECAnet} in exploiting inter-channel relationships of the convolutional feature maps, they have a common property in using global pooling (max and/or average) to summarize information carried by every feature map. The global pooled information is not the optimal feature to infer accurate channel attention~\cite{CBAM,FCANet}. As illustrated in Figure.~\ref{fig:SE_issue}, due to the data dependency of statistical pooling operators (\textit{e.g.} global maximum or global average value), feature maps with similar statistical characteristics can represent distinctive spatial relationships.

% different feature maps yield very similar (non-discriminable) channel descriptors if only global statistical information (\textit{e.g.} global maximum or global average value) is used without considering the spatial relation among the feature maps (\textit{i.e.} channels).

This paper addresses the above issue by modelling CNNs' feature maps as entities of a geographical analysis system in which entities are characterized by their statistical and topological properties simultaneously. The ``First Law of Geography'' states that ``everything is related to everything else, but near things are more related than distant things''~\cite{FL_Geo}. For instance, people are less likely to travel greater distances to visit a store. Therefore, a retailer can determine the best possible locations for a new shop, considering how many people live within a 10-minute drive time from existing stores. Following this premise, the spatial inter-channel relationships of feature maps can tell us about the systematic spatial variation within feature maps, which is needed to address the aforementioned problem related to existing channel-wise attention modules. Here, we introduce Moran's metric from geographical information science~\cite{Moran}, for the first time, into the CNNs domain to analyse the geographical spatial correlation of the channel dependencies. We refer to the new attention unit, which adopts channel-wise spatial autocorrelation of deep feature maps as ``\textit{channel-wise spatially autocorrelated}'' (CSA) attention block.

\begin{figure*}[t]
\centering
	\centering
	\includegraphics[width=0.85\textwidth, scale=0.65]{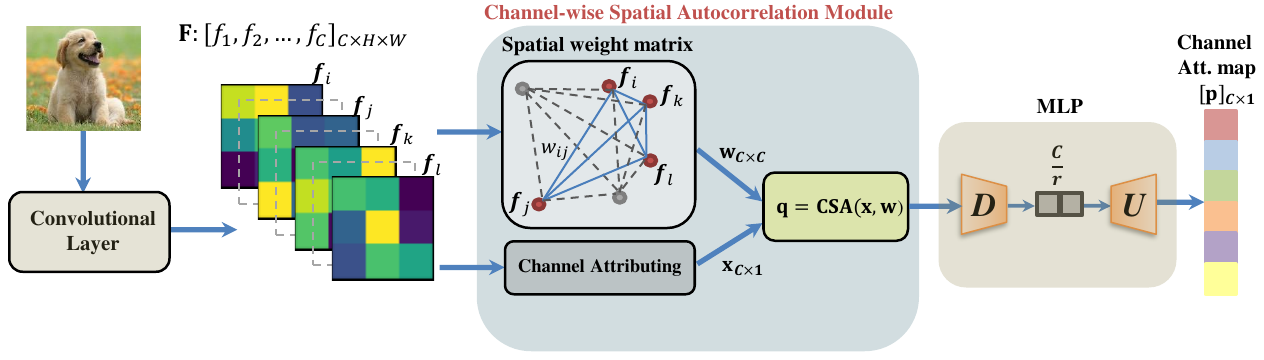}%
	\caption{The proposed CSA framework for modelling channel-wise attention maps. As illustrated, spatial autocorrelation of feature maps is utilized to refine global contextual information. Boxes with different colours demonstrate different values of the computed attention map upon the channel axis. $\mathbf{q}$ refers to the computed spatially autocorrelated channel descriptors. $\mathnormal{D}$ and $\mathnormal{U}$ indicate fully connected layers of the multi-layer perceptron (MLP) for channel reduction and up-sampling, respectively.}
	\label{fig:model_spt}
\end{figure*}

The overview of the proposed CSA framework is illustrated in Figure~\ref{fig:model_spt}. To the best of our knowledge, this is the first time that the concept of geographical spatial analysis is utilized in deep CNNs. The implementation of the proposed CSA is straightforward and imposes negligible parameters overhead to the deep model. Furthermore, it can be easily integrated into baseline deep structures with negligible overhead to the computation cost. Therefore, our proposed CSA can be considered as a beneficial choice for other deep architectures as well.

The extensive experiments on image classification benchmark tasks show that the proposed channel-wise attention paradigm is able to effectively re-calibrate the feature maps by extracting inter-channel spatial dependencies. To verify the generalization capability of the proposed CSA-Net, we demonstrate its effectiveness in object detection and instance segmentation. Further deep investigation studies are also conducted to investigate the characteristics of the proposed CSA.
%-------------------------------------------------------------------------
\section{Related Works}
\label{sec:relatedworks}

% \subsection{CNN Architectures}
% Undoubtedly, much effort has been made to enrich convolutional neural networks (CNNs) feature extraction and representation power. Going deeper and stacking more layers are the most intuitive ways to improve network performance. For example, VGG ~\cite{VGG} show superior performance compared to AlexNet~\cite{AlexNet} which has half fewer layers. However, simply stacking many convolutional layers may lead to vanishing/exploding gradients problems~\cite{glorot_understanding_2010}. Residual nets (ResNets)~\cite{he2016deep,he2016identity} addressed this problem by applying simple identity shortcut connections between stacked layers. Due to adopting identity-based skip-connection, ResNets demonstrate the benefit of very deep networks effectively.

% Apart from increasing the depth of the network, architectures like WideResNet~\cite{WideResNet} mainly concentrate on expanding the width of the network by using a larger number of convolutional filters. Recently, EfficientNet~\cite{EfficientNet} proposed a compound scaling method that uniformly scales the depth, width, and feature resolution of the deep network. On the other hand, Radosavovic \textit{et al.}~\cite{RegNet} proposed regular networks (RegNets) to design network design spaces, where design space is a parametrized set of possible model architectures. RegNets~\cite{RegNet} could outperform EfficientNet~\cite{EfficientNet} in terms of performance and speed.

% % MobileNetV2~\cite{Mobilenetv2},EfficientNet~\cite{EfficientNet},RegNetY-400MF~\cite{RegNet}
% \subsection{Attention Mechanism}
The critical role of a visual attention mechanism in human perception has been shown through a range of studies~\cite{corbetta2002control,zhang2008multiplicative}. The human vision system efficiently and adaptively focuses on salient areas and suppresses other irrelevant information. Attention mechanism has been widely incorporated in computer vision to improve the efficiency of the model~\cite{xu2015show,yohanandan2018saliency}. From the attention mechanism perspective, related works to this paper method can be divided into three main categories: First-order, Second-order and Self-attention techniques. 
% Residual attention network~\cite{ResAtt} introduces a noise-resistant encoder-decoder style module to generate a 3D attention map for refining the intermediate feature maps. However, this attention framework suffers from high computational/parameter overhead. 

\subsection{First-order methods}
First-order statistics (\textit{e.g.} global average pooling) have been widely used in most deep neural networks~\cite{he2016deep,he2016identity,denseNet} to summarize the feature maps representation. Hu~\textit{et al.}~\cite{senet} incorporated the global averaged statistics into a lightweight gating mechanism to exploit inter-channel relationships of the convolutional feature maps. Their squeeze-and-excitation (SE) module measures channel-wise attention by squeezing first-order spatial information. Nevertheless, SENets’ global squeezing function causes two main issues in producing a fine attention map: sub-optimal effect of global average-pooled features and missing spatial axes~\cite{CBAM,BAM}. Following the idea of attention mechanism in SENets, channel-wise attention (also known as feature-based attention) and spatial attention are two main streams to advance attention module in CNNs.

% \vspace{-2.6066pt}
% \setlength{\parskip}{0.15em}
In this regard, the competitive SE (CMPE-SE) network~\cite{CMPE-SE} re-scales the value for each channel by modelling the competition between residual and identity mappings. On the other hand, the bottleneck attention module (BAM)~\cite{BAM} presents a 3D attention map that considers both channel and spatial axes simultaneously. Convolutional block attention module (CBAM)~\cite{CBAM} introduces similar approaches to exploit spatial and channel-wise attention sequentially. Moreover, CBAM~\cite{CBAM} uses global max-pooled features as an extra clue to infer a finer channel attention map. Global-and-local (GALA)~\cite{GALA} combines spatial- and feature-based attentions into a single mask that refines feature representations. The GALA module can be supervised by human feedback to learn what and where to attend. Similarly, batch aware attention module (BA$^2$M)~\cite{BA2M} fuses channel (based on the global average pooling), local spatial, and global spatial attention maps to obtain the sample-wise attention representation (SAR). These SARs are then normalised to derive weights for each sample, capturing varying feature importance across content complexities within the batch.
\setlength{\parskip}{0.15em}

Apart from these techniques, the bridge attention network (BA-Net)~\cite{BANet} integrates previous features into the attention layer using straightforward strategies similar to the SENets~\cite{SENet_j}. The top-down attention module (TDAM)~\cite{TDAM} iteratively generates a ``visual searchlight'' that performs channel and spatial modulation on its inputs and outputs, resulting in more contextually-relevant feature maps at each computation step.
The dense-and-implicit attention network (DIANet)~\cite{dianet} includes a densely connected framework that shares an attention module throughout the network. In particular, its attention mechanism incorporates Long Short-Term Memory (LSTM)~\cite{LSTM}, rather than fully connected layers in SENet~\cite{senet}, to capture long-distance channel-wise dependency. Wang~\textit{et al.}~\cite{ECAnet} developed an efficient channel attention (ECA) network to adaptively determine coverage of local cross-channel interaction in the excitation layer. On the other hand, Ruan \textit{et al.}~\cite{GCT} proposed a parameter-free Gaussian context transformer (GCT), which replaced the fully-connected layers of the SE model and aimed to capture the presumed negative correlation between global contexts and attention activations~\cite{LCT}. Building upon GCT~\cite{GCT}, Chen \textit{et al.}~\cite{SGA} introduced the bilateral asymmetric skewed Gaussian attention (bi-SGA), which combines the skewed and asymmetric properties to further enhance the attention mechanism within GCT~\cite{GCT}.

% \textcolor{darkred}{Despite employing various sophisticated strategies, advanced attention mechanisms often prioritise spatial attention over channel-wise dependencies~\cite{BAM,CBAM,GALA,GCNet,BANet}, or they still suffer from the same drawback inherent in SENets, as depicted in Figure~\ref{fig:SE_issue}: the sub-optimal effect of squeezing global spatial information. This paper highlights the significance of channel-wise spatial autocorrelation in feature maps for inferring finer feature-based attention. Subsequently, we demonstrate that refining pooled features using the proposed channel-wise spatial autocorrelation framework effectively and efficiently addresses the aforementioned issue.}

\subsection{Second order methods}
Higher-order statistics have shown the capability of modelling discriminative image representation~\cite{higherOrd,Cmpact_lin}. The global second-order squeezing methods benefit from covariance matrix representations of images to achieve performance improvement~\cite{GSOP,Is_second,2edOrder_gcov,MultiScale}. Due to the high computation overhead of covariance matrix caused by the high resolution feature maps of the intermediate layers, these methods are mostly plugged at the deep network end. Though several attempts~\cite{Cmpact_lin,higherOrd,kernel_pool,FASON} have tried to alleviate this issue by adopting compact covariance representations, they are only applicable at the deepest layers of deep networks.

In this regard, Gao \textit{et al.}~\cite{GSOP} proposed a global second-order pooling method, which introduces an effective feature aggregation method by considering the pairwise channel correlations of the input tensor. It also utilizes convolutions and non-linear activation to accomplish embedding of the resulting covariance matrix used for scaling the 3D tensor along the channel dimension. However, despite plugging the GSoP~\cite{GSOP} into any layer of deep neural networks, the computation of the generated covariance matrix still imposes high computational costs (FLOPs) on the model. In this paper, unlike other covariance-based second-order approaches, we show that the proposed channel-wise spatial autocorrelation framework offers a more efficient and effective second-order attention mechanism.

\subsection{Self-attention methods}
The proposed method is also related to self-attention mechanisms~\cite{AttAll,SAGAN} from the perspective of modelling long-range dependencies across input feature maps. The self-attention module calculates response at a position in a sequence by weighting the sum of the features at all positions within the same sequence. In general, given the input sequence of a flattened tensor with the shape of ($C\times(HW)$) ($C$, $H$, and $W$ represent the number of channels, height, and width of the input tensor), self-attention algorithms calculate the triplet (key, query, value). Then, the softmax-normalized dot-product of the query with all keys is computed to get the attention scores. Finally, the values are reweighted by the attention scores ~\cite{AttAll,vit_survey,AA,stand-SAN}.

In this regard, Bello \textit{et al.}~\cite{AA} proposed attention augmented (AA) convolutional network by combining convolutional and self-attention processing mechanisms. However, AA~\cite{AA} still is not efficiently applicable throughout the network including early high-resolution layers due to high memory and computational costs. To alleviate this shortcoming, Ramachandran \textit{et al.}~\cite{stand-SAN} introduced stand-alone self-attention (SASA) restricting the scope of self-attention to a local patch (e.g., $7 \times 7$ pixels), rather than applying global self-attention over a whole feature map~\cite{AttAll,AA}. On top of this, Zhao \textit{et al.}~\cite{san-imr} explored two types of vector-based self-attention modules to replace all the spatial convolution layers in the common CNNs and yield the self-attention network (SAN) topology.

Despite the success of self-attention mechanisms for many applications, they suffer from a critical shortcoming of high computational costs, which makes them less practical to be efficiently incorporated with intermediate layers of a deep CNN processing relatively high-resolution feature maps~\cite{vit_survey,AA}. For instance, having a feature map $\mathbf{x} \in \mathbb{R}^{C\times 100\times 100}$ ($C$ represents the number of channels and $C\ll100\times100$), a self-attention module generates a huge spatial attention map with the size of $10000\times10000$ to describe each pixel’s attention score on every other pixel. On the other hand, as discussed in ~\cite{NLN_den,NLNet}, self-attention can be also viewed as a form of the non-local mean and in this sense, some recent methods like gather-excitation (GE)~\cite{GENet}, non-local spatial attention module (NL-SAM)~\cite{SE-NonLocal}, and global context network (GCNet)~\cite{GCNet} effectively model the global spatial context using a non-local network (NLNet)~\cite{NLNet}, which at the same time introduces significant computational expense to the models.

% , and in this sense our
% work bridges self-attention for machine translation to the
% more general class of non-local filtering operations that are
% applicable to image and video problems in computer vision. Though the similarity between our work and self-attention mechanisms, the proposed method treats every feature map independently and extracts autocorrelation relationship along channel dimension while self-attention mechanisms focus on spatial dependencies

\section{Channel-wise Spatially Autocorrelated Attention Module}
\label{sec:CSA}
% \section{Preliminaries: Geographical Spatial Auto-correlation}
\subsection{Geographical Spatial Auto-correlation}
In this paper, we address the global contextual information squeezing problem of channel-wise attention mechanisms~\cite{senet,BAM,CBAM} from a geographical analysis perspective. To bridge the gap between the fields of CNNs and geographical analysis, we briefly review the concept of spatial autocorrelation in geographical science, which is needed to understand our work. 

% Then, we formally present the spatial squeeze-excitation module where features maps are re-calibrated based on their channel-wise spatial statistics.
In geographical modelling, spatial autocorrelation plays an important role to measure spatial dependency of entities based on both entities' locations and values simultaneously. Positive spatial autocorrelation indicates adjacent observations (entities) have similar values, while negative spatial autocorrelation means that nearby observations tend to have contrasting values. In general, there are two types of measures: global measures that summarize the level of spatial autocorrelation with respect to all data points and local measures that provide a value
for each location concerning its neighbourhood. To calculate such measurements, Moran's metric~\cite{Moran} is a technique that is used in geographical analysis. In this paper, to the best of our knowledge, for the first time, we use Moran's measurement to investigate spatial dependency within deep feature maps. Moreover, as
we aim at investigating the spatial properties of each feature map, we focus only on local Moran's indicators~\cite{local_moran}.

Let $\mathbf{F}$ be a set of $C$ observations presented by location vectors $f_i$, $\mathbf{F}=[f_1,f_2,...,f_C]$, and an associated attribute, $\mathbf{x}=[x_1,x_2,...,x_C]$, the local Moran's I metric can be defined as:
\begin{equation}
\bm{\mathit{I}}_{l}(i)= \frac{C (x_{i}-\mu) \sum_{j=1}^{C}v_{ij} (x_{j}-\mu)}{\sum_{i=1}^{C} \sum_{j=1}^{C}v_{ij}\sum_{i=1}^{C} (x_{i}-\mu)^{2}}
\label{eq:orig_moran_local}
\end{equation}
where $\mu$ is the mean value of features upon their attribute. $v_{ij}$ represents the elements of a spatial contiguity matrix $\mathbf{v}=[v_{ij}]_{C\times C}$, in which $v_{ij}$ denotes the degree of closeness or the contiguous relationships between $f_i$ and $f_j$ $(i,j=1,2,...,C)$. 
% The performance of spatial autocorrelation analysis is highly reliant on choosing the right function, $\mathbf{v}=[v_{ij}]_{C\times C}$, to measure pairwise spatial relationships of the feature maps. In this work, we explore two overall approaches: the first is a deterministic method, originally established in spatial autocorrelation literature~\cite{Moran, local_moran, new_moransI}, while the second is grounded in a self-attention mechanism~\cite{AttAll}, which we refer to as the parametric approach.
% \subsubsection{Pairwise spatial contiguity measure }
There are several ways to construct deterministic $\mathbf{v}=[v_{ij}]_{C\times C}$~\cite{four_spatial_wmatrix, new_moransI}. For the sake of simplicity and following~\cite{new_moransI}, we adopt the negative exponential function in this work as:
\begin{align}
\begin{aligned}
v_{ij} &= \begin{cases}
\exp(\frac{-l_{ij}}{\bar l}), & \text{if }i\neq j\\
0 ,& \text{if } i=j.
\end{cases}
\end{aligned}
\label{eq:vij}
\end{align}
where $l_{ij}$\footnote{Although $l_{ij}$ can take any form of distance measurements, it refers to $L_2$ distance between $f_i$ and $f_j$ throughout the paper.} refers to the distance between $f_i$ and $f_j$, and $\bar l$ is the average distance between any two elements.

\subsection{Channel Attributing}
As illustrated in Figure~\ref{fig:model_spt}, we need to adopt the spatial analysis to our channel-wise attention models. This can be done by replacing the notations of observation and system in geographical context with the terms of feature maps and convolutional layer, respectively. In fact, given a convolutional layer (\textit{i.e.} a system) generates a number of observations, which is represented by corresponding feature maps, $f \in \mathbb{R}^{d}$ lies in a $d$-dimensional vector space. Every feature map $f$ is attributed with corresponding channel descriptors (\textit{e.g.} global average value).

% However, the original definition of Moran's I coefficients as formulated in Eq.\ref{eq:orig_moran_local} are not efficient and cannot be integrated with convolutional layers directly. Specifically, due to high number of multidimensional feature maps ($d>>64$) in deep models they impose significant computational overhead and make them impractical. Here, with the aid of matrix calculus and the simplified spatial analysis approach presented by~\cite{new_moransI}, we can define a new efficient spatial channel-wise attention module as illustrated in Figure~\ref{fig:model_spt}. 

In this regard, suppose there is an intermediate feature map $\mathbf{F}=[f_1,f_2,...,f_C] \in \mathbb{R}^{C\times H\times W}$, which is associated with its channel-wise global context features descriptor $\mathbf{x}=[x_1,x_2,...,x_C]\in \mathbb{R}^{C\times 1}$ as:
\begin{align}
\mathbf{x}=\left[ x_{i}=\frac{1}{H\times W}\sum_{h=1}^{H}\sum_{w=1}^{W} f_{i}(h,w)\right]_{C\times 1},
\label{eq:gavg}
\end{align}
where $C$ and $(H,W)$ denote the channel number and spatial dimension of the feature map, $\mathbf{F}$, respectively. $f_{i}(h,w)$ represents the $i$-th channel feature value at position $(h,w)$. Please note that more sophisticated strategies or application-oriented criteria can be applied to obtain the global contextual information descriptor (Eq.~(\ref{eq:gavg})). Here, we use the global average pooling for the sake of simplicity.

\subsection{Channel-wise Spatial Autocorrelation}
As shown in~\cite{new_moransI}, satisfying the further condition of unitary normalization enables a simpler and easier form of Eq.(\ref{eq:orig_moran_local}) for implementation. Therefore, following the technique proposed by Chen~\cite{new_moransI}, the $C$-by-$C$ unitary spatial weight matrix (USWM), $\mathbf{w}$, can be derived from Eq.~(\ref{eq:vij}) as:
\begin{equation}
\mathbf{w}=\left[w_{ij}=\frac{v_{ij}}{\sum_{i=1}^{C}\sum_{j=1}^{C}v_{ij}}\right]_{C\times C}, \sum_{i=1}^{C}\sum_{j=1}^{C}w_{ij}=1
\label{eq:USWM}
\end{equation}
Having the unitary spatial weight matrix, $\mathbf{w}$, Eq.~(\ref{eq:orig_moran_local}) can be simplified as:  
\begin{equation}
 \bm{\mathit{I}}_{l}=[\text{diag}(\mathbf{z}^{t}\mathbf{z}\mathbf{w})]_{C\times 1},
\label{eq:local_new_moran}
\end{equation}
where $\text{diag}(.)$ returns the diagonal elements of a matrix. $\mathbf{z}$ refers to normalised channel-wise global context features as:

\begin{equation}
 \mathbf{z}=\frac{\mathbf{x}^{t}-\mu}{\sigma},
\label{eq:norm_gap}
\end{equation}
where $\mu$ and $\sigma$ denote mean and standard deviation of $\mathbf{x}$, respectively. Symbol ``t'' indicates the transpose of the matrix. 

Since the unitary spatial weights matrix (USWM), $\mathbf{w}=\left[w_{ij}\right]_{C\times C}$, is relatively big ( $C\times C \ge 64 \times 64$), the normalization operation in Eq.~(\ref{eq:USWM}) (i.e., $\frac{v_{ij}}{\sum_{i=1}^{C}\sum_{j=1}^{C}v^{h}_{ij}}$ ) leads to very small values of $w_{ij}$ and consequently results in extremely small values in elements of $\bm{\mathit{I}}_{l}$ (Eq.~(\ref{eq:local_new_moran})). Such too small values in $\bm{\mathit{I}}_{l}$ may exceed the precision of the floating point and be rounded to zero during computation. To avoid this issue, the channel-wise spatial autocorrelation descriptor, $\mathbf{q}$, can be defined as normalised $\bm{\mathit{I}}_{l}$:
\begin{equation}
 \mathbf{q}=\frac{\bm{\mathit{I}}_{l}-\mu_{\bm{\mathit{I}}_{l}}}{\sigma_{\bm{\mathit{I}}_{l}}},
\label{eq:refined_ch_avg}
\end{equation}
where $\mu_{\bm{\mathit{I}}_{l}}$ and $\sigma_{\bm{\mathit{I}}_{l}}$ indicate mean and standard deviation of $\bm{\mathit{I}}_{l}$, respectively.

% Note that the Eq.~(\ref{eq:refined_ch_avg}) can be considered a special case of layer normalization~\cite{LN} operator because of being applied along channel dimension of $\bm{\mathit{I}}_{l}\in \mathbb{R}^{C\times 1}$. As shown in~\cite{LN,U_LN}, it can also result in more stable training of the excitation layers and provide better generalization performance. The new channel descriptor is based on both spatial and contextual inter-dependencies among feature maps.

% \begin{figure}[t]
% \centering
% 	\includegraphics[width=0.85\textwidth, trim=0.0cm 0.0cm 0.0cm 0.0cm, clip=true,scale=0.85]{./figs/S2E-Resblock.pdf}%
% 	\caption{Integration of the proposed CSA framework with a common residual block~\cite{he2016deep}. We apply CSA on the top of every residual block. $\bigotimes$ and $\bigoplus$ represent the element-wise multiplication and summation operations, respectively.}
% 	\label{fig:CSA-SE-resblock}
% \end{figure}

\subsection{Channel Attention Map}
We construct spatially autocorrelated channel-wise gating mechanism by forming
a multi-layer perceptron (MLP) with one hidden layer. The generated channel descriptor from previous step, $\mathbf{q}$, is forwarded to the MLP in which its first fully-connected (FC) layer followed by a non-linearity $\delta$ (ReLU~\cite{relu}) reduces the channel dimension with ratio $r$ (here, we set $r=16$). Then an up-sampling FC layer followed by sigmoid activation, $\alpha$, returns the input channel dimension ($C$). So the channel attention map can be formalized as:
\begin{align}
\mathbf{p}=\alpha(\bm{U}(\delta(\bm{D}(\mathbf{q})))),
\label{eq:ch_att_map}
\end{align}
where $\bm{D}$ and $\bm{U}$ represent the FC layers for channel reduction and up-sampling operators respectively. 

% However, ReLU~\cite{relu} activation has been commonly adopted as $\delta$ function, it may impede our gating mechanism due to disregarding negative values of spatial autocorrelation. In contrast with SENets~\cite{SENet_j} and its derived topologies like CBAM~\cite{CBAM,BAM}, here, negative spatial autocorrelation can be important as it shows dissimilarity of neighboring channels. On the other hand other benchmark non-linear functions like To 
% \begin{align}
% \delta=\text{tanh}(y)(1+\frac{|y|}{1+e^{-y}}),
% \label{eq:delta}
% \end{align}
% where $y=\bm{D}(\mathbf{q})$.
% \section{Integrating with convolutional layer}

% Then the Eq.~\ref{eq:orig_moran}, Moran's index, can be expressed in the quadratic form:
% \begin{equation}
% \bm{\mathit{I}}_{l}_{global}= \mathbf{Z}^{t}\mathbf{W}\mathbf{Z},
% \label{eq:new_moran}
% \end{equation}

\subsection{Integration with Deep Networks}\label{subsec:integerat_cnn}
The proposed CSA framework can be easily integrated with deep architectures. The given output feature maps $\mathbf{F}\in \mathbb{R}^{C\times H\times W}$ of a convolutional unit is adaptively refined by the spatially autocorrelated channel-wise attention map, $\mathbf{p}\in \mathbb{R}^{C\times 1\times 1}$. The overall process can be summarized as:
\begin{align}
\begin{split}
\mathbf{F}= \mathbf{p} \otimes \mathbf{F},\\
% &\mathbf{F}=\mathbf{F} \oplus \acute{\mathbf{F}},
\end{split}
\label{eq:integr_net}
\end{align}
where $\otimes$ denotes element-wise multiplication. In this work, we integrate the CSA module into the final convolution layer within each resolution scale of the deep network. 
% To enhance efficiency, particularly for high-resolution images such as those from ImageNet and MS-COCO with input dimensions ($\ge 224 \times 224$), feature maps larger than $8 \times 8$ are downsampled by a factor of $r=2$ before being inputted into the CSA module.

\subsection{Implementation Details}
We use stochastic gradient descent (SGD) with a similar optimization configuration as~\cite{he2016identity} for the training of all models. In particular, we adopt Nesterov momentum \cite{sutskever2013importance} with a momentum weight of 0.9, and $10^{-4}$ as the weight decay. 
% For convolutional units we use $3\times3$  and $1 \times 1$ kernel sizes. Moreover, one pixel zero paddings are applied to each side to keep the feature map size constant.

On ImageNet, we train our models for 100 epochs with a mini-batch size of 128. For this experiment, the initial learning rate is set to 0.1 and is decreased by a factor of 10 at epochs 30, 60, and 90, respectively. For MS-COCO-2017 object detection and instance image segmentation tasks, models are trained for 24 epochs. All models are developed using PyTorch version 1.10 and trained on two NVIDIA GeForce V100 GPU cards with CUDA/cuDNN 11.0 toolkits.

% All models are trained with a mini-batch size of 128 for 300 epochs on CIFAR-10/100. The initial learning rate is set to 0.1, divided by 10 at $50\%$ and $75\%$ of the total number of training epochs. On ImageNet, we train our models for 100 epochs with a mini-batch size of 64. For this experiment, the initial learning rate is set to 0.1 and is decreased by a factor of 10 at epochs 30, 60, and 90, respectively. For MS-COCO-2017 object detection and instance image segmentation tasks, models are trained for 24 epochs. Different to CIFAR-10/100, the ImageNet and MS-COCO images have  much higher dimensions, their feature maps are downsampled by a factor of 2 before feeding to the CSA model. All models are developed using PyTorch version 1.10 and trained on two NVIDIA GeForce V100 GPU cards with CUDA/cuDNN 11.0 toolkits.

\section{Experimental Results and Discussion}
\label{sec:exp-discuss}

The effectiveness of the proposed CSA model is evaluated across different deep architectures as well as different tasks. We use ImagNet-1K~\cite{ImageNet} for image classification. Further, MS-COCO-2017~\cite{coco} is utilized for object detection and instance segmentation tasks. 

\subsection{ImageNet Image Classification}
\begin{table}[t]
\centering
\caption{The top-1 and top-5 classification error rates (\%) on ImageNet validation set. Single-crop validation errors are reported. Note that only channel-wise attention maps are used in the CBAM$^{*}$'s attention mechanism~\cite{CBAM}. The first and second best results are indicated in boldface and underlined-face, respectively.}
\vspace*{-2mm}
\resizebox{\linewidth}{!}{%
\begin{tabular}{l|c|c|c|c}
\hline
\multicolumn{1}{c|}{Method} & GFLOPs &Params. & \multicolumn{1}{l|}{top-1 err.} & \multicolumn{1}{l}{top-5 err.} \\ \hline\hline
% \multicolumn{5}{c}{ResNet-50~\cite{he2016deep}}\\\hline\hline
ResNet-50~\cite{he2016deep} &3.86&25.6M&24.70&7.80\\ 
+SE~\cite{senet,SENet_j}&3.87&28.1M&23.14&6.70\\
+CBAM$^{*}$~\cite{CBAM}&3.87&28.1M&22.80&6.52\\
+CBAM~\cite{CBAM}&3.87&28.1M&22.66&6.32\\
+BAM~\cite{BAM}&3.94&25.9M&24.02&7.18\\
+GALA~\cite{GALA}&4.1&29.4M&22.73&6.35\\
+DIA~\cite{dianet}&-&28.4M&22.76&-\\
+NL-SAM~\cite{SE-NonLocal}& 7.03&32.5M&24.45&7.51\\
+AA~\cite{AA}& 8.30&25.8M&22.30&6.20\\
+SASA~\cite{stand-SAN}& 7.20&18.0M&22.40&-\\
% \textcolor{darkred}{SAN (pairwise)~\cite{san-imr}}&\textcolor{darkred}{SAN19~\cite{san-imr}}& 3.80&18.0M&23.10&6.60\\
+GSoP~\cite{GSOP}&6.56&28.2M&22.49&6.24\\
+ECA~\cite{ECAnet}&4.13&25.5M&22.55&6.32\\
+FCA~\cite{FCANet}&4.13&28.4M&21.43&5.90\\
+GCT~\cite{GCT}&4.12&25.6M&22.45& 6.29\\
+BA-Net~\cite{BANet}&4.13&28.7M&\textbf{21.15}& \textbf{5.72}\\
+bi-SGA~\cite{SGA}&4.12&25.6M&22.36& 6.3\\
+CSA (Ours)&4.12&26.2M&\uline{21.41}&\textbf{5.72}\\[0.7ex]\hline\hline
% \multicolumn{5}{c}{ResNet-101~\cite{he2016deep}}\\\hline\hline
ResNet-101~\cite{he2016deep} &7.58& 44.5M&23.3&6.88\\
+SE~\cite{senet,SENet_j}&7.60& 49.3M&22.35 &6.19\\
% CBAM$^{*}$~\cite{CBAM}&7.58& 49.3M&22.85 &6.51\\
+BAM~\cite{BAM}&7.65&49.9M&22.40&6.29\\
+AA~\cite{AA}&16.10&45.4M&21.30&5.62\\ 
+GSoP~\cite{GSOP}&12.12&48.9M&22.08&6.05\\
+ECA~\cite{ECAnet}&7.86&44.5M&21.35&5.66\\
+FCA~\cite{FCANet}&7.86&49.3M&\textbf{20.37}&5.37\\
+GCT~\cite{GCT}&7.86&44.5M&21.15& 5.59\\
+BA-Net~\cite{BANet}&7.87&50.5M&\uline{20.97}& \textbf{5.17}\\
+bi-SGA~\cite{SGA}&7.86&44.5M&21.09& 5.55\\
+CSA (Ours)&7.84&45.2M&\uline{20.97} &\uline{5.30}\\[0.7ex] \hline
\end{tabular}
}
\label{ImgNeT_Table_result}
\vspace*{-4mm}
\end{table}
In order to demonstrate CSA's effectiveness on large-scale datasets and in a rigorous condition, we evaluate CSA with two baseline networks, \textit{i.e.} ResNet-50 and ResNet-101~\cite{he2016deep}, and their attention-based variations on the ImageNet classification task~\cite{AlexNet}. ImageNet consists of 1.28 million images for training and 50,000 images for validation, respectively, from 1,000 classes. Following~\cite{he2016deep,he2016identity}, we adopt similar data augmentation for training data alongside single-cropping of $224\times 224$. We also report classification error rates on the validation set. To the best of our knowledge, a few attention mechanisms~\cite{FCANet,GSOP} have considered and tried to address the data dependency problem of global pooling. In addition to GSoP~\cite{GSOP}, Qin~\textit{et al.}~\cite{FCANet} presented a frequency channel attention (FCA) mechanism which generalizes the squeezing of the channel attention mechanism in the frequency domain. Other methods either leverage spatial attention (it is not channel-wise) and max-pooling (BAM~\cite{BAM} and CBAM~\cite{CBAM}), or focus on enhancing the excitation layer (GALA~\cite{GALA}, NL-SAM~\cite{SE-NonLocal}, DIA~\cite{dianet}, ECA~\cite{ECAnet}, GCT~\cite{GCT}, and bi-SGA~\cite{SGA}) to improve SE baseline, while they still use the same squeezing method as that of SENets~\cite{senet,SENet_j}.
% Since the main purpose of the proposed method is addressing the existing problem of global pooling in existing attention modules, for a fair comparison, counterparts that tackle similar issues as us are required.

Table~\ref{ImgNeT_Table_result} shows the mean top-1 and top-5 classification error rates obtained by these models. It can be seen that networks equipped with the CSA module significantly outperform the ResNet-50 and ResNet-101 baselines and SE benchmarks. Moreover, the models with CSA outperform those counterparts that benefit from spatial resolution attention map (BAM~\cite{BAM}, CBAM~\cite{CBAM}, GALA~\cite{GALA}). These results demonstrate the effectiveness of the proposed spatial autocorrelation analysis of feature maps in generating richer channel descriptors.

The CSA also achieves lower error rates than the second order method GSoP~\cite{GSOP} and the non-local/self-attention methods NL-SAM~\cite{SE-NonLocal}, AA~\cite{AA}, and SASA~\cite{stand-SAN} with about half computational expense (GFLOPs). The result shows the effectiveness and yet the efficiency of CSA against benchmarks. Though CSA obtains competitive performance against FCA~\cite{FCANet} and BA-Net~\cite{BANet}, our further evaluations on object detection (see Table~\ref{coco_result}) and instance segmentation (see Table~\ref{InsSeg_result}) tasks reveal that the CSA shows superior generalization ability on the three different tasks.
\subsection{MS-COCO Object Detection}
\begin{table}[t]
\centering
\caption{Object detection mAP(\%) on the MS-COCO 2017 validation set. ResNet-50~\cite{he2016deep} is used as the backbone network for all models. The best results are indicated in boldface.}
\resizebox{\linewidth}{!}{%

\begin{tabular}{l|c|c|c|c|c|c}
\hline
\multicolumn{1}{c|}{Method} &  \multicolumn{1}{l}{mAP@[.5,.95]} & \multicolumn{1}{l|}{mAP@.5} & \multicolumn{1}{l|}{mAP@.75}&\multicolumn{1}{l}{mAP$_S$}&\multicolumn{1}{l}{mAP$_M$}&\multicolumn{1}{l}{mAP$_L$} \\ \hline\hline
Faster-RCNN~\cite{Faster-RCNN}&  36.4 & 58.2 & 39.2&21.8&40.0&46.2\\ 
+SE~\cite{SENet_j}& 37.7&60.1&40.9&22.9&41.9&48.2\\
+AA~\cite{AA}& 39.4&61.0&42.3&-&-&-\\
% \textcolor{darkred}{SASA~\cite{stand-SAN}}& &  & 38.2&60.8&41.5&23.0&42.5&47.8\\
+GSoP~\cite{GSOP}& 39.4&60.5&42.9&23.2&43.1&49.8\\
+ECA ~\cite{ECAnet}&38.0&60.6&40.9&23.4&42.1&48.0\\
+FCA~\cite{FCANet}&39.0&\textbf{61.1}&42.3&23.7&42.8&49.6\\
+GCT~\cite{GCT}&38.9&60.4&42.3&22.8&43.1&49.7\\
+BA-Net~\cite{BANet}&39.5&61.3&43.0&\textbf{24.5}&43.2&50.6\\
+bi-SGA~\cite{SGA}& 39.2&60.5&42.5&23.2&43.1&49.8\\
+CSA (Ours)& \textbf{39.7}&\textbf{61.1}&\textbf{43.1}&23.4&\textbf{43.5}&\textbf{51.3}\\\hline
%-----------------------------------------------------
Mask-RCNN~\cite{mask_rcnn}&  37.2 & 58.9 & 40.3&22.2&40.7&48.0\\ 
+SE~\cite{SENet_j}& 38.7&60.9&42.1&23.4&42.7&50.0\\
+GSoP~\cite{GSOP}&  39.9&60.4&43.5&23.7&42.9&52.3\\
+ECA ~\cite{ECAnet}& 39.0&61.3&42.1&24.2&42.8&49.9\\
+FCA~\cite{FCANet}&  40.3&\textbf{62.0}&44.1&\textbf{25.2}&43.9&52.0\\
+GCT~\cite{GCT}& 39.4&60.8&42.9&23.6&43.3&50.7\\
+BA-Net~\cite{BANet}& 40.5&61.7&44.2&24.5&\textbf{44.3}&52.1\\
+bi-SGA~\cite{SGA}& 38.6&61.0&43.0&23.6&43.5&50.8\\
+CSA (Ours)& \textbf{40.5}&61.6&\textbf{44.4}&24.3&44.1&\textbf{53.0}\\\hline
\end{tabular}
}
\label{coco_result}
\end{table}

% \subsection{MS-COCO Object Detection}
% \begin{table*}[t]
% \centering
% \caption{Object detection mAP(\%) on the 40K MS-COCO validation set. The best results are indicated in boldface.}
% \begin{tabular}{l|c|c|c|c}
% \hline
% \multicolumn{1}{c|}{Backbone} & Detector & \multicolumn{1}{l|}{mAP@.5} & \multicolumn{1}{l|}{mAP@.75} & \multicolumn{1}{l}{mAP@[.5,.95]} \\ \hline\hline
% ResNet-50~\cite{he2016deep}&Faster-RCNN~\cite{Faster-RCNN}& 46.2 & 28.1 & 25.1\\ 
% SE-ResNet-50~\cite{senet}&Faster-RCNN~\cite{Faster-RCNN}& 46.8&N/A&26.4\\
% % SE-CliqueNet& 9.50M&45.36&21.36\\ 
% CSA-ResNet-50 (Ours)&Faster-RCNN~\cite{Faster-RCNN}&\textbf{47.3}&\textbf{30.0}&\textbf{28.3}\\\hline
% ResNet-101~\cite{he2016deep}& Faster-RCNN~\cite{Faster-RCNN} &48.4&30.7&27.2\\
% SE-ResNet-101~\cite{senet}& Faster-RCNN~\cite{Faster-RCNN}&49.2 &N/A&27.9\\
% CSA-ResNet-101 (Ours)& Faster-RCNN~\cite{Faster-RCNN}&\textbf{50.1} &\textbf{32.4} & \textbf{29.8}\\\hline
% \end{tabular}
% \label{coco_result}
% \end{table*}
In addition to the image classification task, we also use the Microsoft (MS)-COCO 2017 dataset~\cite{coco2017} to evaluate the generalization capability of the proposed CSA module on object detection tasks. MS-COCO is a large-scale object detection, segmentation, key-point detection, and captioning dataset. In this work, we conduct object detection on this dataset. Microsoft-COCO dataset splits into two sets of 115K (``2017 train'') and 5K (``2017 val'') images for training and validating purposes, respectively. Following~\cite{ECAnet,FCANet}, and \cite{SENet_j}, we adopt Faster-RCNN~\cite{Faster-RCNN} and Mask-RCNN~\cite{mask_rcnn} as the detection models and ImageNet pre-trained ResNet-50 as the backbone network. Here, following~\cite{SENet_j}, we plug CSA into base architecture to evaluate its benefit. The MS-COCO's standard evaluation metric mAP over different IoU with thresholds from 0.5 to 0.95 as well as mAP on small (mAP$_S$), medium-size (mAP$_M$) and large (mAP$_L$) ground truth objects are reported, respectively. We use SE~\cite{SENet_j}, AA~\cite{AA}, GSoP~\cite{GSOP}, ECA~\cite{ECAnet}, FCA~\cite{FCANet}, GCT~\cite{GCT}, BA-Net~\cite{BANet}, and bi-SGA~\cite{SGA} for comparison. As shown in Table~\ref{coco_result}, CSA can outperform its counterparts on most evaluation metrics demonstrating the generalization ability of the proposed CSA module on different tasks.

\begin{table}[t]
% \scalebox{0.5}
\centering
\caption{Instance segmentation mAP(\%) using Mask-RCNN~\cite{mask_rcnn} on the MS-COCO 2017 validation set . ResNet-50~\cite{he2016deep} is used as the backbone network for all models. The best results are indicated in boldface.}
\resizebox{\columnwidth}{!}{%
\begin{tabular}{l|c|c|c}
\hline
\multicolumn{1}{c|}{Method} & \multicolumn{1}{l|}{mAP@[.5,.95]} &\multicolumn{1}{l|}{mAP@.5} & \multicolumn{1}{l}{mAP@.75}  \\ \hline\hline
Mask-RCNN~\cite{mask_rcnn}& 34.1 & 55.5 & 36.2\\ 
+SE~\cite{SENet_j}&35.4 & 57.4 & 37.8\\
+GSoP~\cite{GSOP}& 36.0 & 57.5 & 38.2\\
+ECA~\cite{ECAnet}&35.6 & 58.1 & 37.7\\
+FCA~\cite{FCANet}& 36.2 & 58.6 & 38.6\\
+GCT~\cite{GCT}&35.7 & 57.6 & 38.0\\
+BA-Net~\cite{BANet}& \textbf{36.6} & 58.7 & 38.6\\
+bi-SGA~\cite{SGA}& 35.9 & 57.6 & 38.1\\
+CSA (Ours)& 36.5&\textbf{58.8}& \textbf{38.8}\\\hline
\end{tabular}}
\label{InsSeg_result}
\end{table}
\subsection{MS-COCO Instance Segmentation}
We further conduct experiments on the instance segmentation task. In particular, we present instance segmentation results of our CSA module using Mask R-CNN on MS-COCO 2017. The standard evaluation metric mAP with the threshold from 0.5 to 0.95 is reported in Table~\ref{InsSeg_result}. The results show that superior performance of the proposed CSA can be achieved for most of the different settings.

% \begin{table}[t]
% \centering
% \caption{Ablation study of the hybrid spatial contiguity measure ($\mathbf{v}^{h}$). Results attained by CSA-ResNet-110 on CIFAR-100 image classification task. $l_{ij}$ and $l$ refer to the pairwise and the average $L_2$ distance of feature maps, respectively. The best results are indicated in boldface.}
% \begin{tabular}{l|c|c}
% \hline
% \multicolumn{1}{c|}{$\mathbf{v}^{h}=[v^{h}_{ij}]$} & Type & \multicolumn{1}{l}{top-1 err.}  \\ \hline\hline
% % \textcolor{darkred}{$l_{ij}^{-b} (b=-1)$}&$L_2$&26.4\\ \hline
% $\exp(\frac{-l_{ij}}{\bar l})$&$L_2$&25.9\\ \hline
% $\frac{<f_{i},f_{j}>}{\parallel f_{i} \parallel\parallel f_{j} \parallel}$&Directional ($\theta$)&25.7\\\hline
% $\frac{<f_{i},f_{j}>}{\parallel f_{i} \parallel\parallel f_{j} \parallel }\times\exp(\frac{-l_{ij}}{\bar l})$&Hybrid&\textbf{25.0}\\\hline
% \end{tabular}
% \label{table:vij_abl}
% \end{table}

\subsection{Analysis and Discussion}
% In this section, the computational overhead of the proposed CSA module is evaluated. Moreover, the effect of the proposed spatially autocorrelated channel descriptor ($\mathbf{q}$) and final deep features are investigated. 
% \subsubsection{Hybrid spatial contiguity measure} 
% The degree of closeness relationships between feature maps is an important factor in obtaining their spatial autocorrelation. Here, we investigate the effectiveness of our proposed technique to define a hybrid pairwise spatial contiguity measure ($\mathbf{v}^{h}=[v^{h}_{ij}]_{C\times C}$). To this end, three experiments based on CSA-ResNet-110 are conducted as shown in Table~\ref{table:vij_abl}. As it can be seen from Table~\ref{table:vij_abl}, the new $\mathbf{v}^{h}$ (Eq.~\ref{eq:vij_2}) achieves almost 1\% lower top-1 validation error than that of other types of measurements.

\begin{table}[t]
% \scalebox{0.5}
\centering
\caption{Comparison of the training and inference time per batch between the baseline ResNet-50, the proposed CSA model and the SE~\cite{SENet_j} counterpart integrated into a ResNet-50 network and trained on ImageNet. All methods' input resolution and batch size are $224 \times 224$ and 128, respectively.}
\resizebox{\columnwidth}{!}{%
\begin{tabular}{l|c|c|c}
\hline
\multicolumn{1}{c|}{Method} & \multicolumn{1}{l|}{training time} &\multicolumn{1}{l|}{inference time} & \multicolumn{1}{l}{top-1 err.}  \\ \hline\hline
ResNet-50~\cite{he2016deep}& 0.134s & 0.133s & 24.70\\ 
+SE~\cite{SENet_j}&0.426s & 0.290s & 23.14\\
+CSA (Ours)& 0.508s&0.281s& \textbf{21.36}\\\hline
\end{tabular}}
\label{table:compute_exp}
\end{table}

% \begin{table}[t]
% % \scalebox{0.5}
% \centering
% \caption{Ablation analysis of the downsampling ratio ($r$). Results attained by CSA-ResNet-50 on ImageNet image classification task.}
% \resizebox{\columnwidth}{!}{%
% \begin{tabular}{l|c|c|c}
% \hline
% \multicolumn{1}{c|}{Ratio $r$} & \multicolumn{1}{l|}{GFLOPs} & \multicolumn{1}{l|}{training speed per batch}& \multicolumn{1}{l}{top-1 err.}  \\ \hline\hline
% 1& 34.1 & & 21.35\\ 
% 2&35.4 & & 21.36\\
% 3& 36.0 & & 21.95\\
% 4& 36.0 & & 22.48\\\hline
% \end{tabular}
% }
% \label{table:down-r}
% \end{table}

% \subsubsection{Downsampling ratio ($r$)}
% The downsampling ratio $r$ introduced in Section~\ref{subsec:integerat_cnn} serves as a crucial hyperparameter that adjusts the model's CSA block capacity and computational cost. We conducted experiments using CSA-ResNet-50 to explore this relationship across various $r$ values. The comparison in Table~\ref{table:down-r} indicates that setting $r = 2$ achieved a favourable balance between accuracy and complexity, and as a result, we adopted this value for all experiments.

\subsubsection{Computational Expense}
Involving more computation is the inevitable cost of adopting channel-wise spatially autocorrelated attention units. Table~\ref{table:compute_exp} compares the time efficiency between the baseline ResNet-50, the proposed CSA model and the SE counterpart integrated into a ResNet-50 network. Under common settings for vision applications, utilising the ResNet-50 architecture and a batch size of 128 with input images at a resolution of $224\times 224$, the CSA exhibits marginal increments in training time per batch—specifically, an increase of 0.082s. Notably, the proposed CSA shows a slightly quicker inference time compared to the SE method.

\begin{figure*}
  \centering
  % First row of figures
  \begin{subfigure}[b]{1\textwidth}
    \centering
    \includegraphics[trim=2.25cm 0.cm .2cm 1.25cm, clip, width=0.245\linewidth]{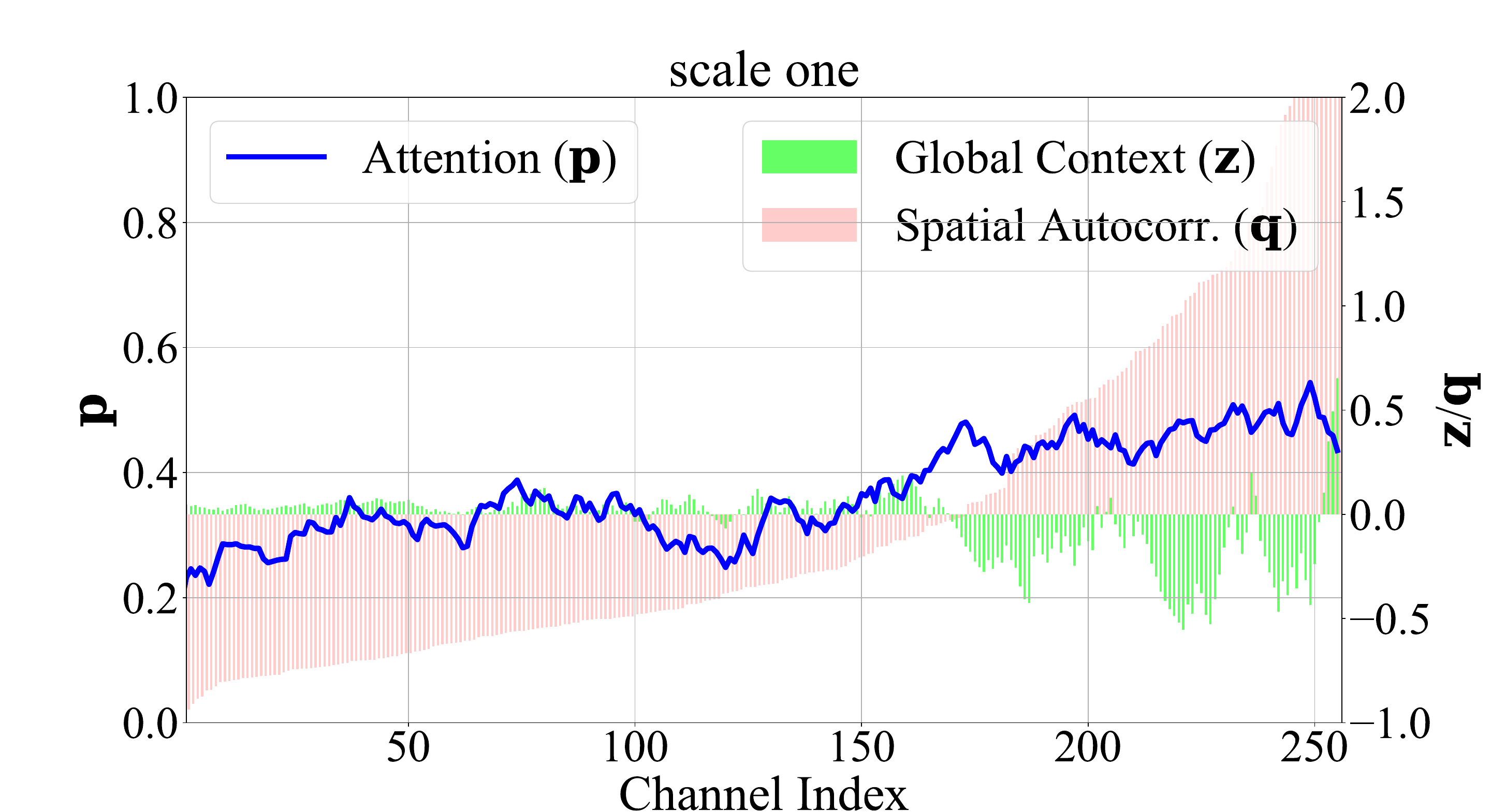}
      \hfill
    \centering
    \includegraphics[trim=2.25cm 0.cm .2cm 1.25cm, clip, width=0.245\linewidth]{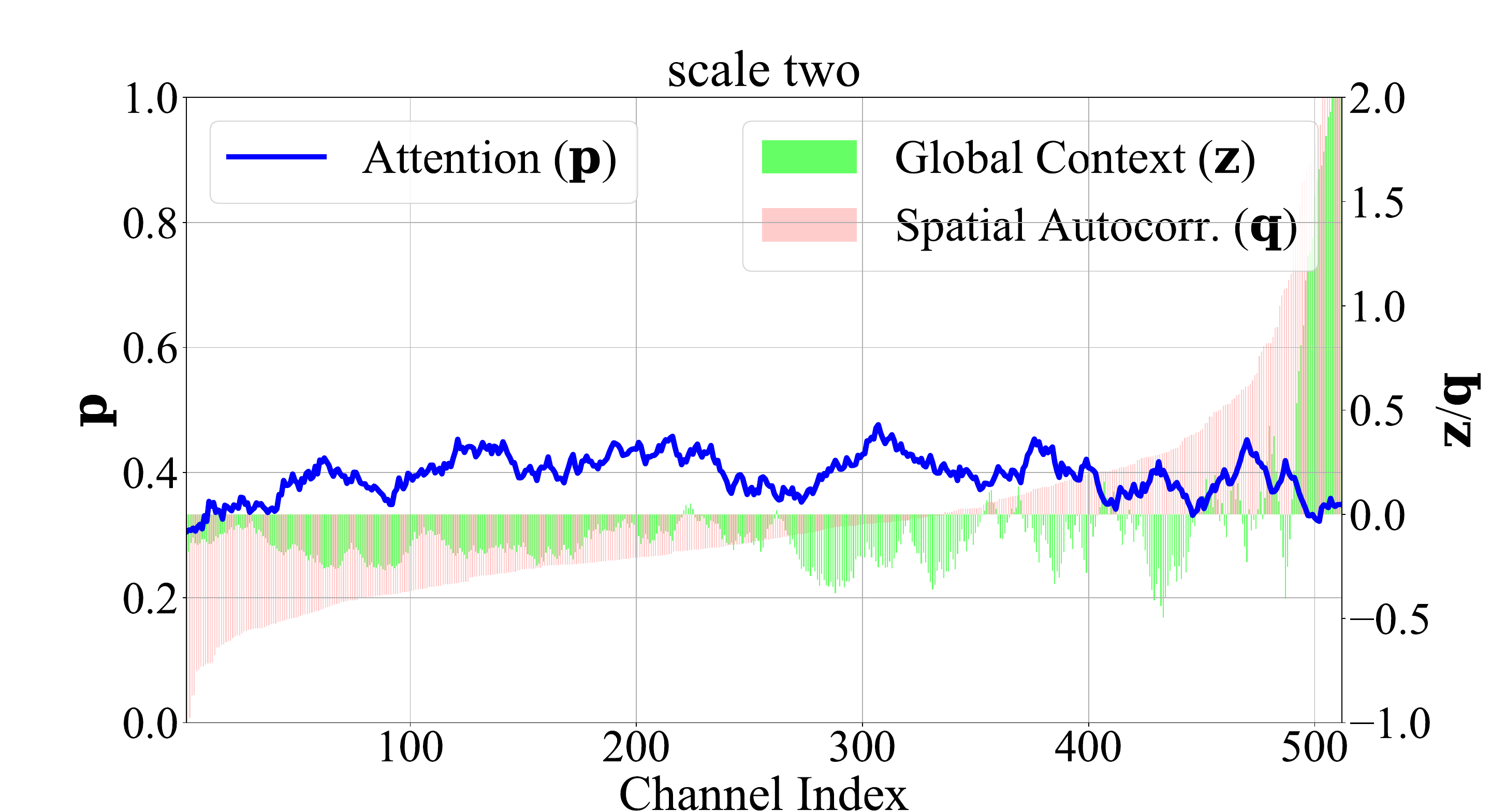}
      \hfill
    \centering
  \includegraphics[trim=2.25cm 0.cm .2cm 1.25cm, clip, width=0.245\linewidth]{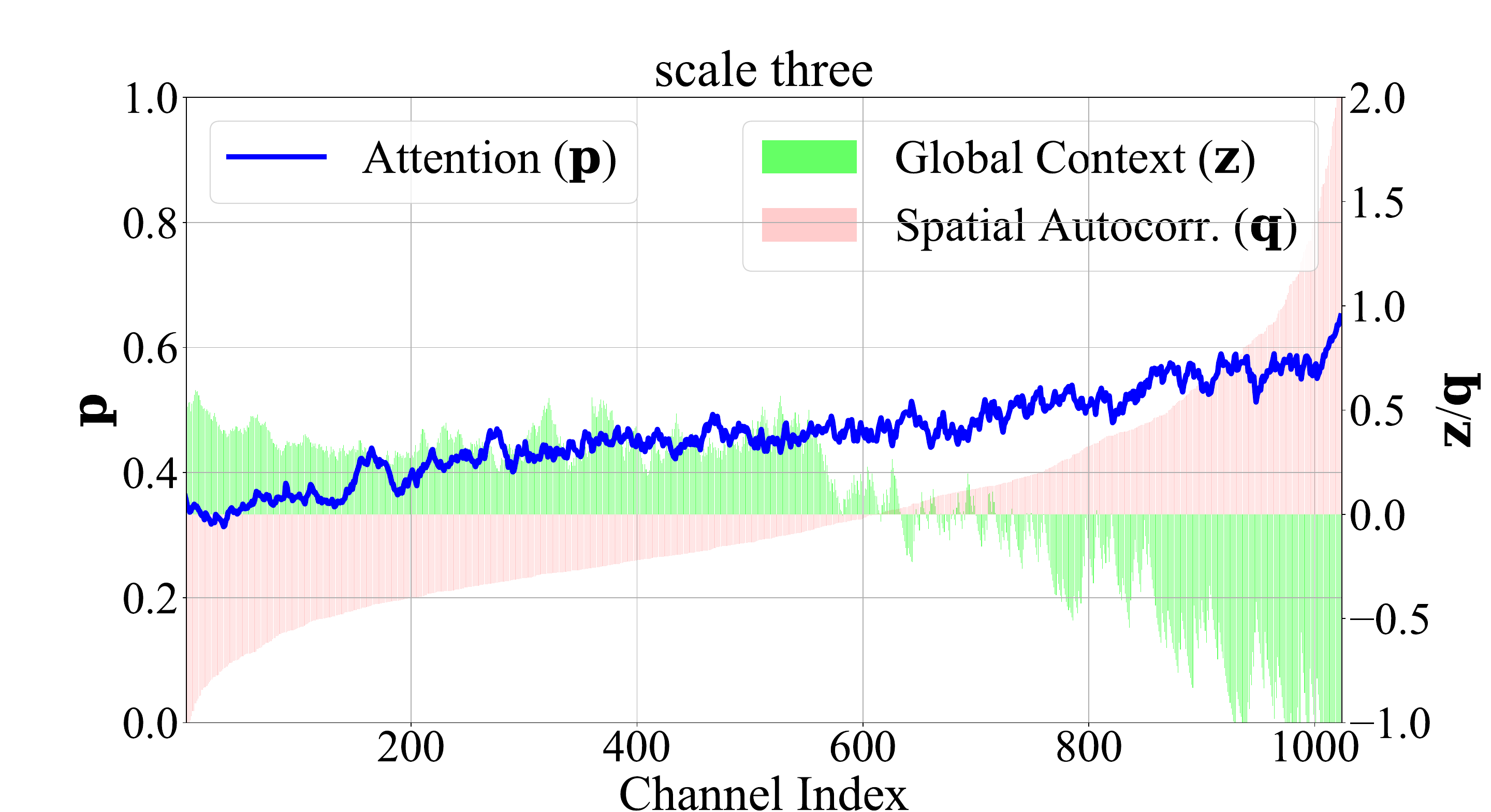}
    \hfill
    \centering
  \includegraphics[trim=2.25cm 0.cm .2cm 1.25cm, clip, width=0.245\linewidth]{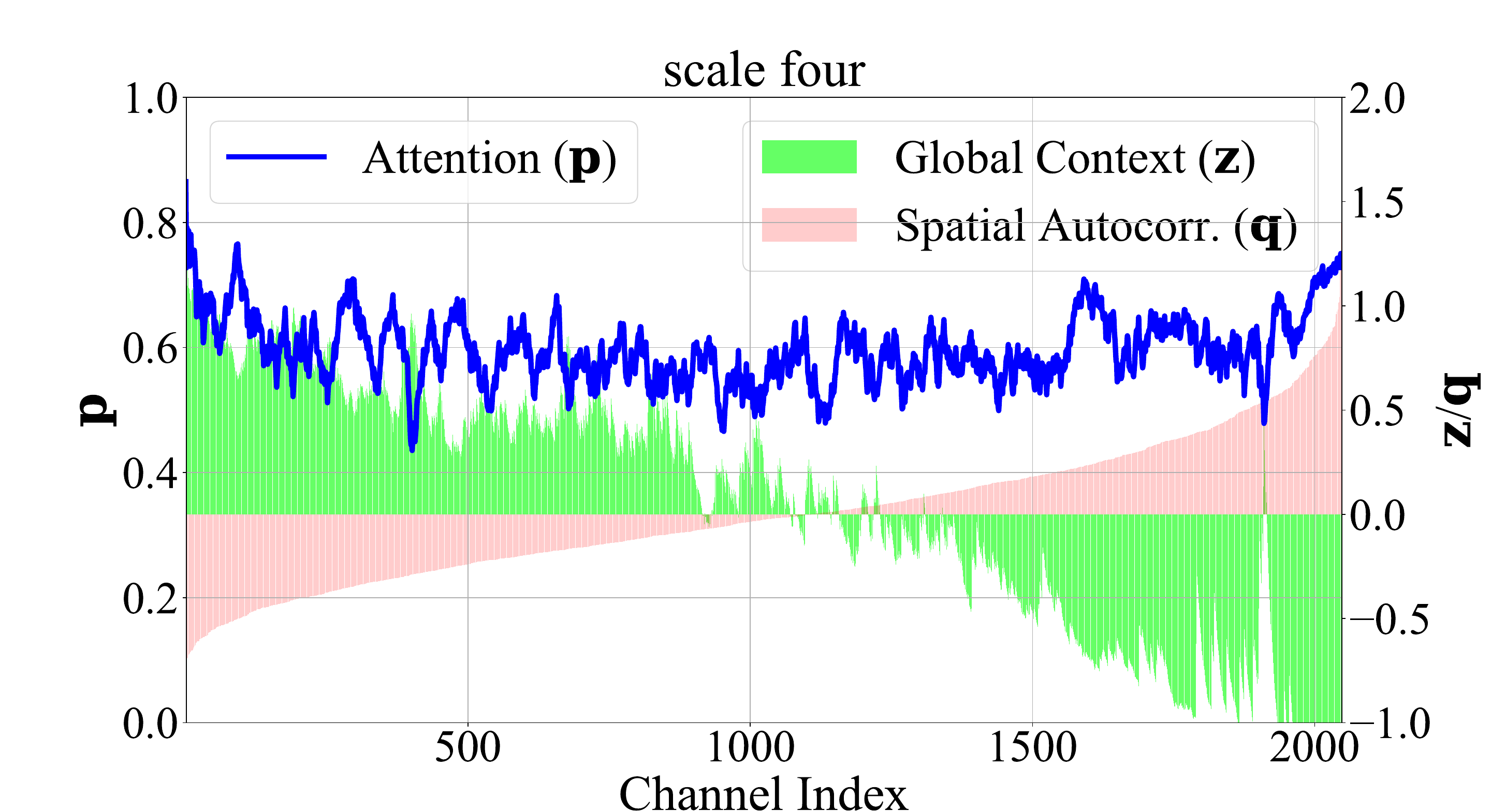}  
\caption{ SE~\cite{SENet_j}}
  \end{subfigure}

  \vspace{1em} % Add vertical space between rows

  % Second row of figures
  \begin{subfigure}[b]{1\textwidth}
    \centering
    \includegraphics[trim=2.25cm 0.cm .2cm 1.25cm, clip, width=0.245\linewidth]{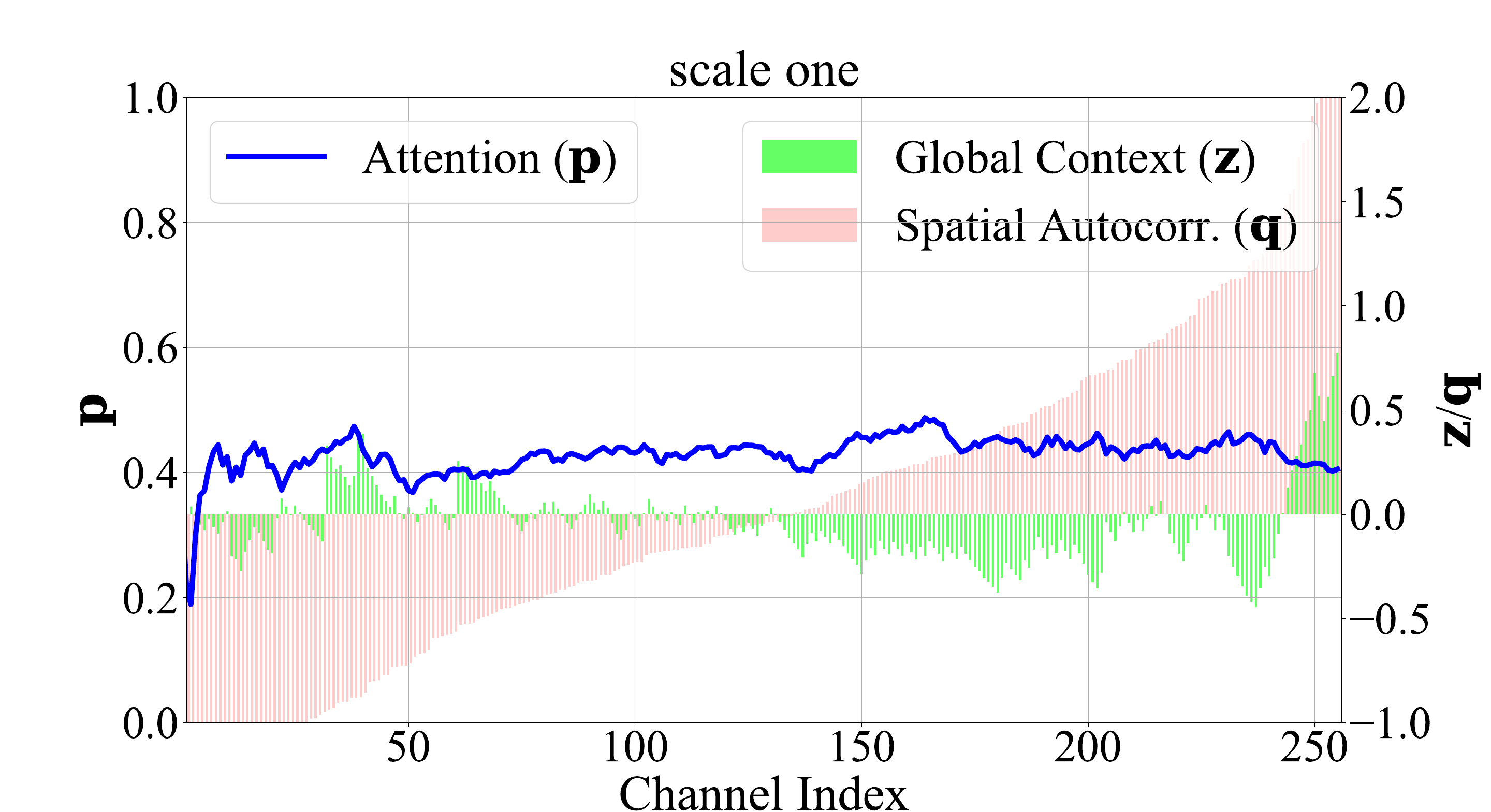}    
  \hfill
    \centering
    \includegraphics[trim=2.25cm 0.cm .2cm 1.25cm, clip, width=0.245\linewidth]{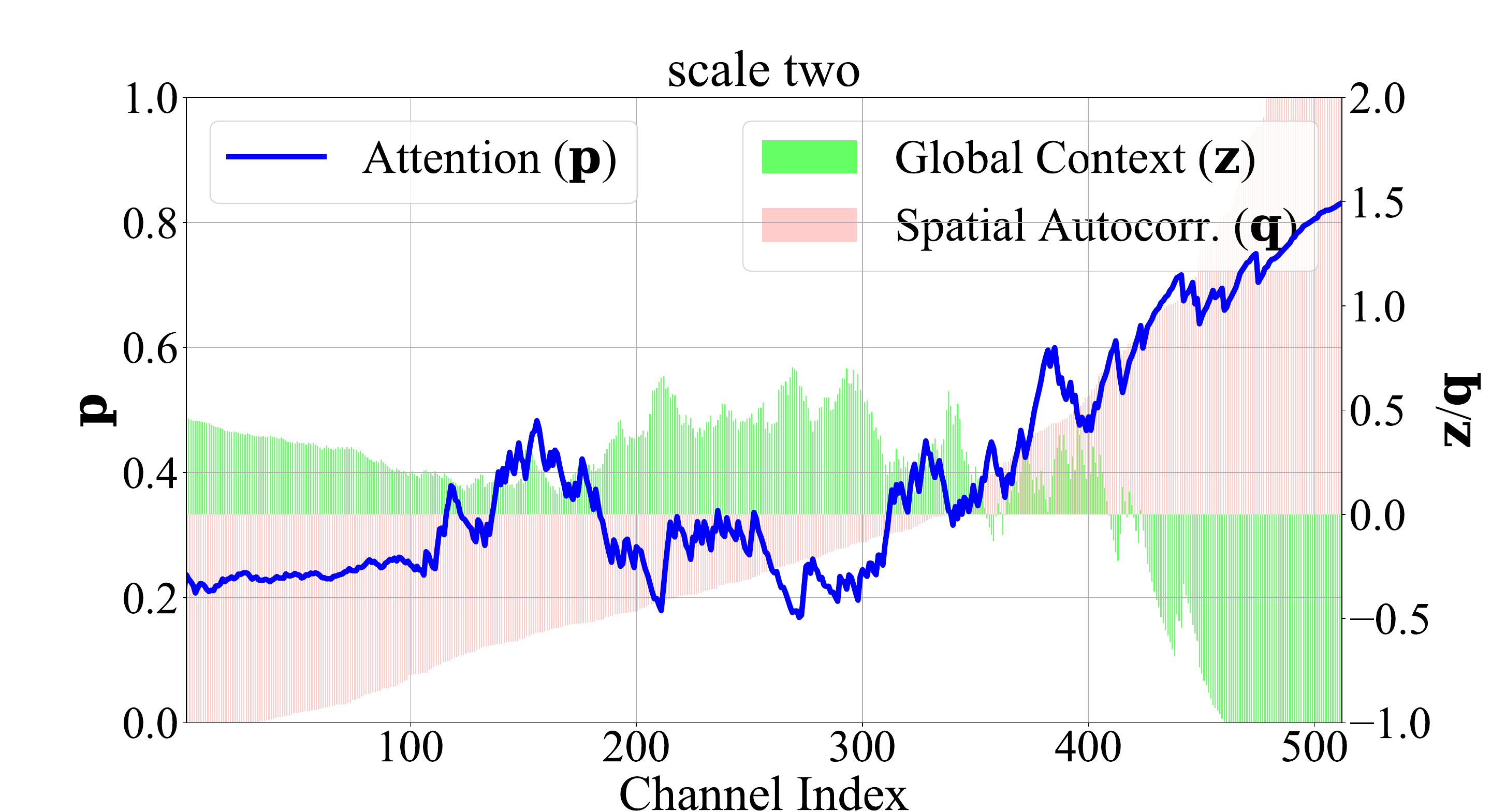}
  \hfill
    \centering
    \includegraphics[trim=2.25cm 0.cm .2cm 1.25cm, clip, width=0.245\linewidth]{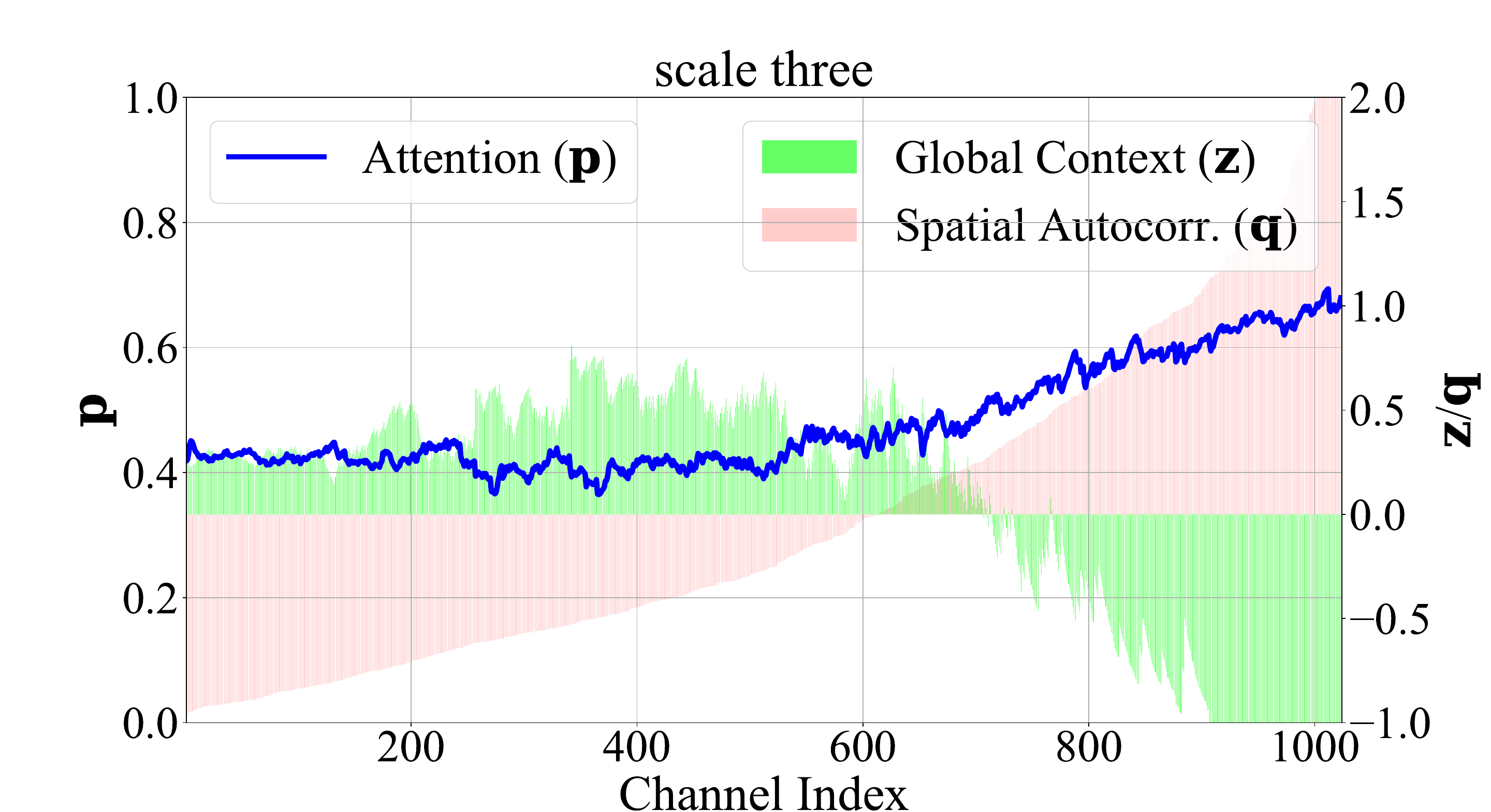}
 \hfill
    \centering
    \includegraphics[trim=2.25cm 0.cm .2cm 1.25cm, clip, width=0.245\linewidth]{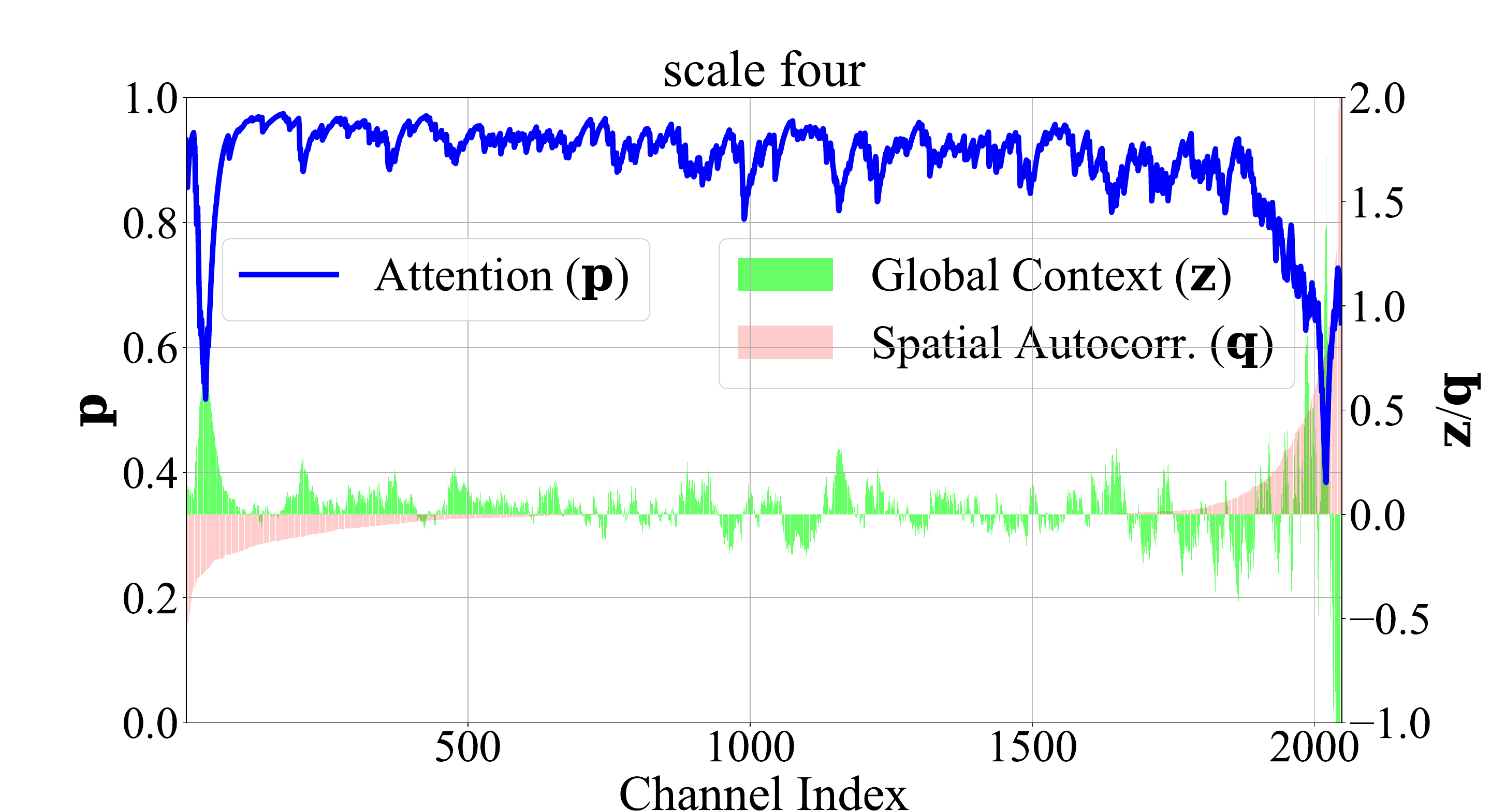}    
\caption{CSA (Ours)}

  \end{subfigure}
 
  % \vspace{1em} % Add vertical space between rows

\caption{Visualising the relationship between averaged channel-wise global context  ($\mathbf{z}$), channel-wise spatial autocorrelation  ($\mathbf{q}$), and corresponding channel attention map ($\mathbf{p}$) produced by the proposed CSA and SE~\cite{SENet_j} attention mechanisms in the last convolutional block at each resolution scale of ResNet-50 on the ImageNet validation set. Please note: the attention values for SE~\cite{SENet_j} are based on their global average pooling, while the proposed CSA is based on channels' spatial autocorrelation. For better presentation, the graphs are smoothed by the exponential moving average with a factor of $0.3$.}
\label{fig:spt_img_scale}
\end{figure*}

\begin{figure}[hbpt]
     \centering
    \includegraphics[trim=.1cm .02cm .50cm 0.0cm, clip=true,scale=0.58]{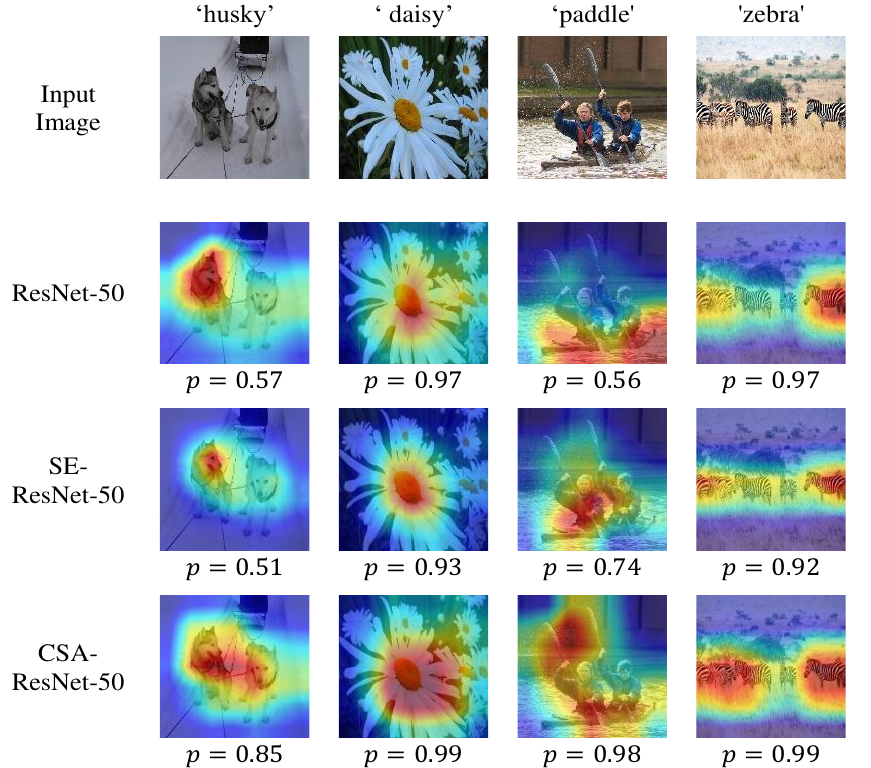}
        \caption{Visual comparisons of Grad-CAM~\cite{grad-cam} analysis generated by the last convolutional outputs layer in ResNet-50, SE-ResNet-50~\cite{SENet_j}, and CSA-ResNet-50 (Ours) on ImageNet. The target class label is shown on the top of each input image, and $p$ indicates the probability score of each model for the target class.}
        \label{fig:grad_cam}
\end{figure}

\subsubsection{Spatially Autocorrelated Channel Descriptor}
Though the effectiveness of the proposed CSA module has been proved empirically, we make a deeper investigation to provide a better understanding of its behaviour. To this end, we examine the relationship between the channel-wise global context (\textbf{$\mathbf{z}$}), channel-wise spatial autocorrelation ($\mathbf{q}$) and the corresponding channel attention map ($\mathbf{p}$). In particular, we develop two distinct models by incorporating SE~\cite{SENet_j}, and the proposed CSA attention modules into the ResNet-50 network. These models are subsequently trained on the ImageNet dataset. Then, we compute and analyse the averaged global context, averaged spatial autocorrelation features and corresponding attention values across the 1000 classes in the ImageNet validation set. Please note that the calculation of spatial autocorrelation measure does not involve any learnable parameters. Therefore, it is applicable to and can be computed for other trained networks such as the SE model as well.

Figure~\ref{fig:spt_img_scale} illustrates the results from the last convolutional layer at all resolution scales of ResNet-50. The averaged spatial autocorrelation features are sorted in ascending order to facilitate the observation. As shown in~\cite{interpretable_ML, interpreting_deep}, low-level features are generated at the early stages. Notably, our CSA attention module maintains a nearly uniform attention distribution among channels for the first resolution scale, indicating that it preserves all the extracted low-level features generated at the early stage. This approach differs from SE, which prioritises channels based on their global contextual values.

In scales two and three, our proposed CSA shows a positive correlation between spatial autocorrelation and attention values. This implies CSA's capability to leverage spatial dependency in intermediate stages of deep networks, giving higher attention to features with higher spatial autocorrelation for the process of subsequent scales. Similar trends are observed in SE, however, with a weaker positive correlation at scale three. 

Figure~\ref{fig:spt_img_scale} reveals an interesting trend among two models in the last scale, where the spatial autocorrelation among their feature maps tends to be lower, indicating reduced spatial correlation during the generation of final features. Notably, the proposed CSA method demonstrates the lowest channel-wise spatial autocorrelation, approaching zero for most channels. This suggests that CSA's final features are almost spatially uncorrelated, which proves by~\cite{discriminative_uncorrelated, uncorrelated_clustering} to be more discriminative features.

\subsubsection{Network Visualization}
Besides analyzing the above-mentioned characteristics of the spatially autocorrelated channel descriptor, we investigate the effectiveness of the proposed CSA from another perspective. For this purpose, we apply the Gradient-weighted Class Activation Mapping (Grad-CAM)~\cite{grad-cam} to qualitatively analyze different networks using the ImageNet validation set. Grad-CAM is a class-discriminative localization technique generally used to visualize particular parts of the image that influence the whole model's decision for an assigned class label. In particular, it uses the gradients of any target concept with respect to the latest activation map to produce a coarse localization map highlighting the crucial regions in the image for predicting the concept~\cite{grad-cam}. As shown in~\cite{grad-cam}, observing the crucial areas that the network considers for predicting a class provides us with a qualitative visual intuition of how the model performs in the effective use of features.

Figure~\ref{fig:grad_cam} illustrates the visualization comparison of the Grad-CAM masks generated by the proposed CSA and SE attention modules integrated with the baseline ResNet-50. It can be seen that the CSA network’s results cover the class-discriminative regions better than other methods, which demonstrates its superior ability in aggregating features and exploiting information from the target object. Moreover, CSA network produces the highest target classes' probability scores.
\section{Conclusion}
\label{sec:conclusion}
In this paper, we present a novel channel-wise spatially autocorrelated (CSA) attention module. For the first time, with the light of geographical spatial analysis, CSA utilizes an effective spatial autocorrelation schema to efficiently exploit spatial inter-dependencies among channels of feature maps. This property enables the CSA-Nets to improve the representational capacity of the network by adaptively re-calibrating channel-wise spatially autocorrelated features. Our extensive experiments verify that the CSA-Nets are able to present a better performance compared to recently advanced attention paradigms.

Apart from the demonstrated novelty and effectiveness of the proposed method, several directions can be considered as future works. One possibility is enhancing the CSA attention module with parametric embedding methods to effectively and yet efficiently model spatial contiguity matrix and feature correlations. Another possible prospective study can be investigating the CSA’s potential as a channel descriptor for other attention mechanisms and applications beyond its current scope. For instance, it could be evaluated as a replacement for conventional global average pooling, unlocking new possibilities for attention-based schemes.
{
    \small
    \bibliographystyle{ieeenat_fullname}
    \bibliography{main}
}

\clearpage
\setcounter{page}{1}
\maketitlesupplementary
% Paper ID: 7960\\

\section{Introduction}
\label{sec:supp-intro}

Due to the page limit in the main text, we provide high-resolution graphs for Subsection~\textcolor{red}{4.4.2} (refer to Figure~\textcolor{red}{3} in the main text) and additional Grad-Cam~\cite{grad-cam} visual comparisons for Subsection~\textcolor{red}{4.4.3} (refer to Figure~\textcolor{red}{4} in the main text) in the following sections. To maintain context between the main text and the provided figures, we reiterate the relevant text here as well.

\begin{figure*}
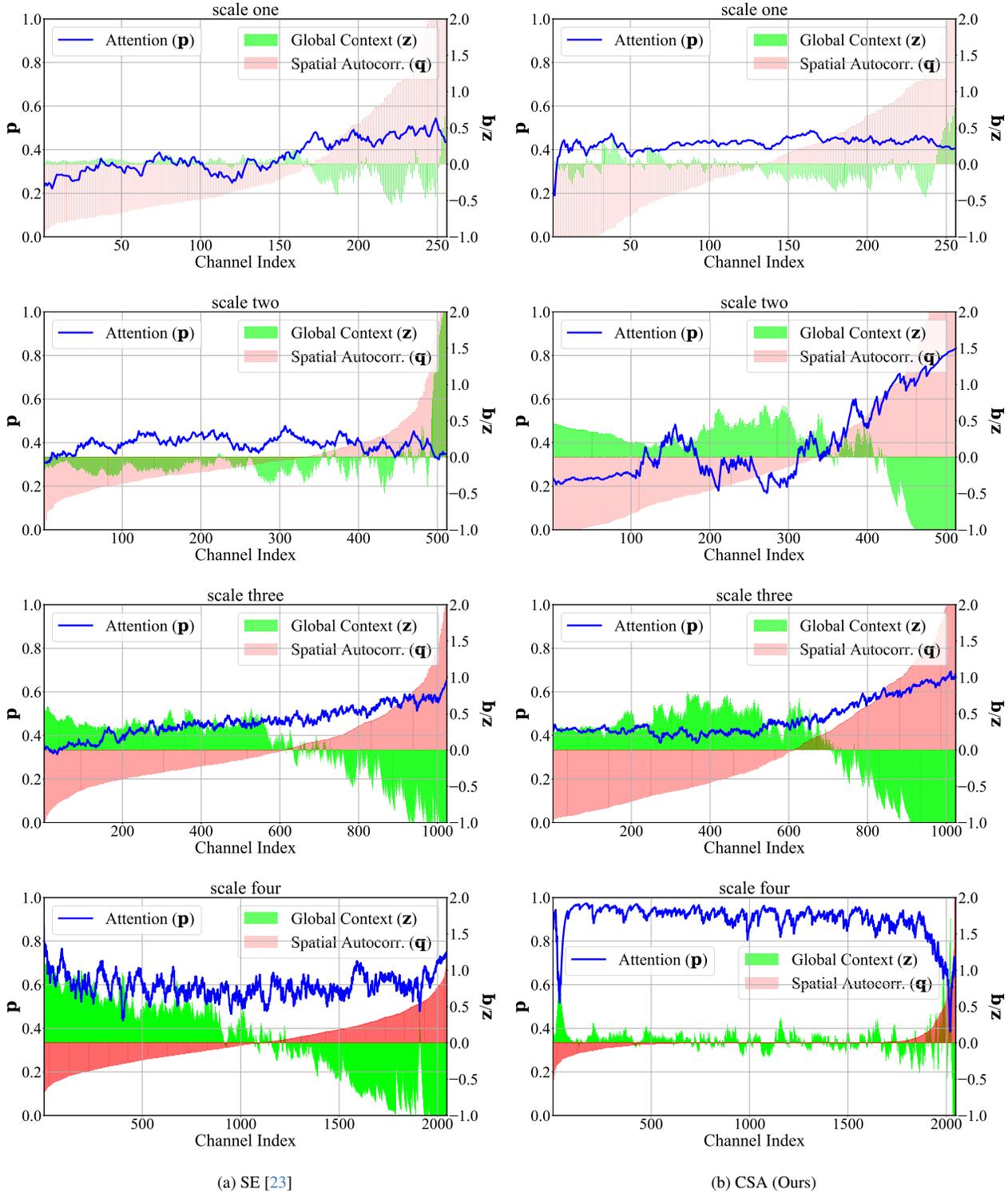

  % \centering
  % First column of figures
  \begin{subfigure}[b]{0.48\textwidth}
    \includegraphics[trim=2.25cm 0.cm .2cm 1.25cm, clip, width=0.98\linewidth]{./figs/ch-analysis/se/se_s1b3.pdf}
    \hfill
    \vspace{1em}
    \hfill
    \includegraphics[trim=2.25cm 0.cm .2cm 1.25cm, clip, width=0.98\linewidth]{./figs/ch-analysis/se/se_s2b4.pdf}
      \hfill
    \vspace{1em}
  \includegraphics[trim=2.25cm 0.cm .2cm 1.25cm, clip, width=0.98\linewidth]{./figs/ch-analysis/se/se_s3b6.pdf}
    \hfill
     \vspace{1em}
  \includegraphics[trim=2.25cm 0.cm .2cm 1.25cm, clip, width=0.98\linewidth]{./figs/ch-analysis/se/se_s4b3.pdf}
\vspace{1em}
\caption{ SE~\cite{SENet_j}}
  \end{subfigure}
% \hfill
     % \vspace{1em}
  % Second column of figures
  \begin{subfigure}[b]{0.48\textwidth}
    \includegraphics[trim=2.25cm 0.cm .2cm 1.25cm, clip, width=0.98\linewidth]{./figs/ch-analysis/csa/csa_s1b3.pdf}
  \hfill
    \vspace{1em}
    \includegraphics[trim=2.25cm 0.cm .2cm 1.25cm, clip, width=0.98\linewidth]{./figs/ch-analysis/csa/csa_s2b4.pdf}
  \hfill
   \vspace{1em}  
    \includegraphics[trim=2.25cm 0.cm .2cm 1.25cm, clip, width=0.98\linewidth]{./figs/ch-analysis/csa/csa_s3b6.pdf}
 \hfill
    \vspace{1em}
    \includegraphics[trim=2.25cm 0.cm .2cm 1.25cm, clip, width=0.98\linewidth]{./figs/ch-analysis/csa/csa_s4b3.pdf}
    \vspace{1em}
\caption{CSA (Ours)}
  \end{subfigure}
  % \vspace{1em} % Add vertical space between rows

\caption{Visualising the relationship between averaged channel-wise global context  ($\mathbf{z}$), channel-wise spatial autocorrelation  ($\mathbf{q}$), and corresponding channel attention map ($\mathbf{p}$) produced by the proposed CSA and SE~\cite{SENet_j} attention mechanisms in the last convolutional block at each resolution scale of ResNet-50 on the ImageNet validation set. Please note: the attention values for SE~\cite{SENet_j} are based on their global average pooling, while the proposed CSA is based on channels' spatial autocorrelation. For better presentation, the graphs are smoothed by the exponential moving average with a factor of $0.3$.}
\label{fig:spt_img_scale}
\end{figure*}

\section{Spatially Autocorrelated Channel Descriptor}\label{sec:SPA}
Though the effectiveness of the proposed CSA module has been proved empirically, we make a deeper investigation to provide a better understanding of its behaviour. To this end, we examine the relationship between the channel-wise global context (\textbf{$\mathbf{z}$}), channel-wise spatial autocorrelation ($\mathbf{q}$) and the corresponding channel attention map ($\mathbf{p}$). In particular, we develop two distinct models by incorporating SE~\cite{SENet_j}, and the proposed CSA attention modules into the ResNet-50 network. These models are subsequently trained on the ImageNet dataset. Then, we compute and analyse the averaged global context, averaged spatial autocorrelation features and corresponding attention values across the 1000 classes in the ImageNet validation set. Please note that the calculation of spatial autocorrelation measure does not involve any learnable parameters. Therefore, it is applicable to and can be computed for other trained networks such as the SE model as well.

Figure~\ref{fig:spt_img_scale} illustrates the results from the last convolutional layer at all resolution scales of ResNet-50. The averaged spatial autocorrelation features are sorted in ascending order to facilitate the observation. As shown in~\cite{interpretable_ML, interpreting_deep}, low-level features are generated at the early stages. Notably, our CSA attention module maintains a nearly uniform attention distribution among channels for the first resolution scale, indicating that it preserves all the extracted low-level features generated at the early stage. This approach differs from SE, which prioritises channels based on their global contextual values.

In scales two and three, our proposed CSA shows a positive correlation between spatial autocorrelation and attention values. This implies CSA's capability to leverage spatial dependency in intermediate stages of deep networks, giving higher attention to features with higher spatial autocorrelation for the process of subsequent scales. Similar trends are observed in SE, however, with a weaker positive correlation at scale three. 

Figure~\ref{fig:spt_img_scale} reveals an interesting trend among two models in the last scale, where the spatial autocorrelation among their feature maps tends to be lower, indicating reduced spatial correlation during the generation of final features. Notably, the proposed CSA method demonstrates the lowest channel-wise spatial autocorrelation, approaching zero for most channels. This suggests that CSA's final features are almost spatially uncorrelated, which proves by~\cite{discriminative_uncorrelated, uncorrelated_clustering} to be more discriminative features.

\begin{figure*}[hbpt]
     \centering
    \includegraphics[trim=.1cm .02cm .50cm 0.0cm, clip=true,scale=0.75]{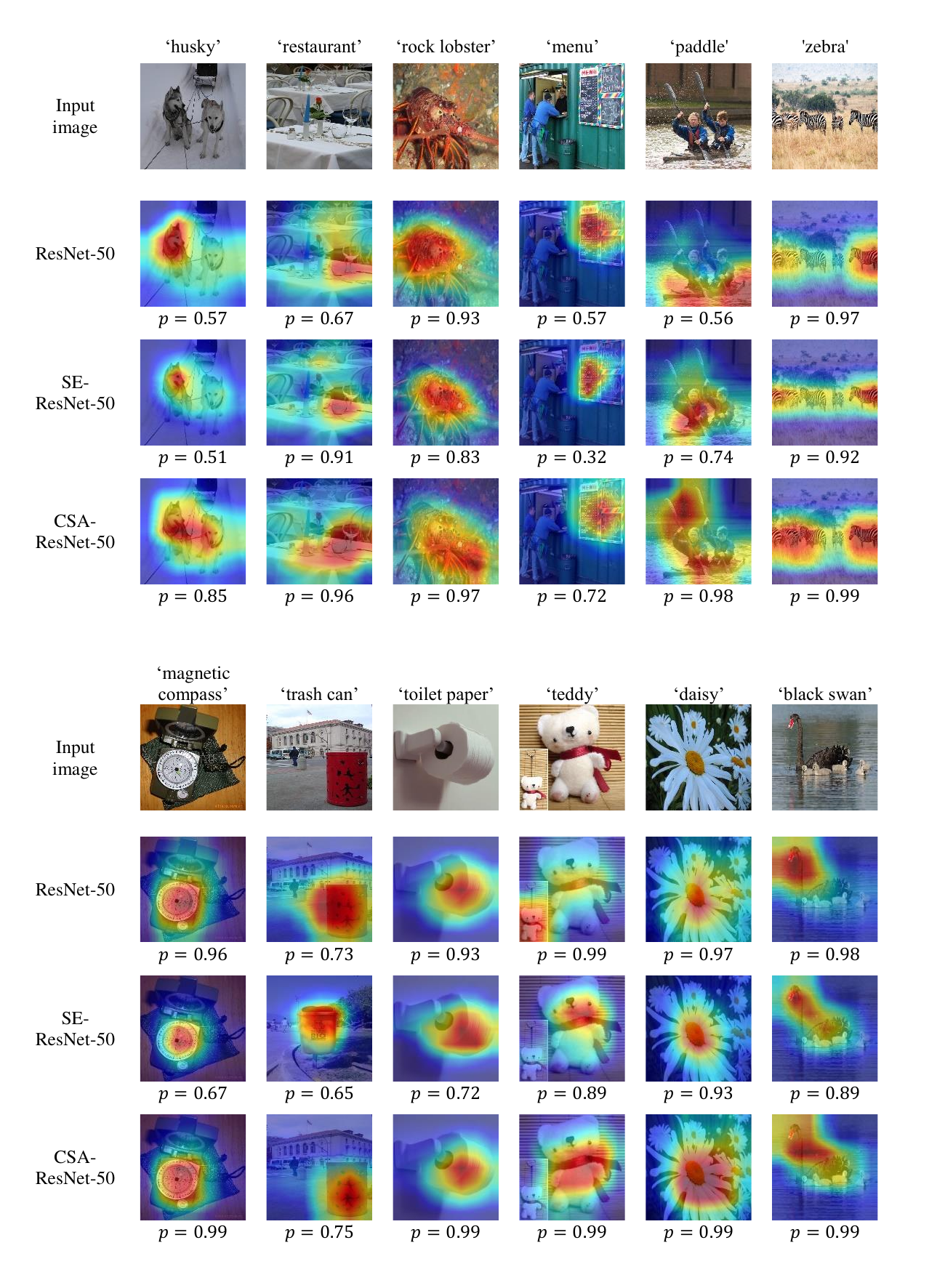}
        \caption{Visual comparisons of Grad-CAM~\cite{grad-cam} analysis generated by the last convolutional outputs layer in ResNet-50, SE-ResNet-50~\cite{SENet_j}, and CSA-ResNet-50 (Ours) on ImageNet. The target class label is shown on the top of each input image, and $p$ indicates the probability score of each model for the target class.}
        \label{fig:grad_cam}
\end{figure*}

\section{Network Visualization}\label{sec:net-viz}
Besides analyzing the above-mentioned characteristics of the spatially autocorrelated channel descriptor, we investigate the effectiveness of the proposed CSA from another perspective. For this purpose, we apply the Gradient-weighted Class Activation Mapping (Grad-CAM)~\cite{grad-cam} to qualitatively analyze different networks using the ImageNet validation set. Grad-CAM is a class-discriminative localization technique generally used to visualize particular parts of the image that influence the whole model's decision for an assigned class label. In particular, it uses the gradients of any target concept with respect to the latest activation map to produce a coarse localization map highlighting the crucial regions in the image for predicting the concept~\cite{grad-cam}. As shown in~\cite{grad-cam}, observing the crucial areas that the network considers for predicting a class provides us with a qualitative visual intuition of how the model performs in the effective use of features.

Figure~\ref{fig:grad_cam} illustrates the visualization comparison of the Grad-CAM masks generated by the proposed CSA and SE attention modules integrated with the baseline ResNet-50. It can be seen that the CSA network’s results cover the class-discriminative regions better than other methods, which demonstrates its superior ability in aggregating features and exploiting information from the target object. Moreover, CSA network produces the highest target classes' probability scores.

\end{document}